\documentclass{article}

\usepackage[preprint]{neurips_2026}

\usepackage[utf8]{inputenc} 
\usepackage[T1]{fontenc}    
\usepackage{hyperref}       
\usepackage{url}            
\usepackage{booktabs}       
\usepackage{amsfonts}       
\usepackage{nicefrac}       
\usepackage{microtype}      
\usepackage{xcolor}         
\usepackage{graphicx}
\usepackage{amsmath}
\usepackage{subcaption}
\usepackage{tikz}
\usetikzlibrary{arrows.meta, positioning}
\usepackage{enumitem}

\newtheorem{theorem}{Theorem}
\newtheorem{proposition}[theorem]{Proposition}
\newtheorem{assumption}[theorem]{Assumption}

\newtheorem{corollary}[theorem]{Corollary}

\title{On What We Can Learn from Low-Resolution Data}
\author{%
  Theresa Dahl Frehr \textsuperscript{1}\thanks{This work was supported by William Demant Fonden (24-5326)} \quad
  Niels Henrik Pontoppidan\textsuperscript{2} \quad
  Hiba Nassar\textsuperscript{1} \quad
  Tommy Sonne Alstrøm\textsuperscript{1} \\[0.5em]
  \textsuperscript{1}Technical University of Denmark \quad
  \textsuperscript{2}Eriksholm Research Center \\
  \texttt{\{tdafr, hibna, tsal\}@dtu.dk} \quad
  \texttt{npon@eriksholm.dk}
}

\begin{document}

\maketitle

%

\begin{abstract}
Artificial intelligence systems typically rely on large, centrally collected datasets, a premise that does not hold in many real-world domains such as healthcare and public institutions. In these settings, data sharing is often constrained by storage, privacy, or resource limitations. For example, small wearable devices may lack the bandwidth or energy capacity needed to store and transmit high-resolution data, leading to aggregation during data collection and thus a loss of information. As a result, datasets collected from different sources may consist of a mixture of high- and low-resolution samples. Despite the prevalence of this setting, it remains unclear how informative low-resolution data is when models are ultimately evaluated on high-resolution inputs. We provide a theoretical analysis based on the Kullback–Leibler divergence that characterises how the influence of a datapoint changes with resolution, and derive bounds that relate the relative contribution of high- and low-resolution observations to the information lost under downsampling. We validate these results in simulation and further show empirically, using both a vision transformer and a convolutional neural network, that low-resolution data can improve performance when high-resolution data is scarce, across controlled benchmarks and a naturally heterogeneous medical dataset. 

\end{abstract}

\section{Introduction}\label{sec:introduction}
Large-scale artificial intelligence (AI) systems rely on massive datasets, substantial storage, computation, and data-transfer resources. In many real-world settings, however, such as healthcare systems, public institutions, and distributed data ecosystems, these resources are limited, and sharing raw data across institutions is often constrained by privacy, regulatory, security, or storage requirements \citep{elmestari-PreservingData-2024, zerka-SystematicReview-2020, fang-DecentralisedCollaborative-2024, pati-PrivacyPreservation-2024}. This makes it difficult to assemble the large centralised datasets on which modern AI systems depend.

\begin{figure}
    \centering
    \includegraphics{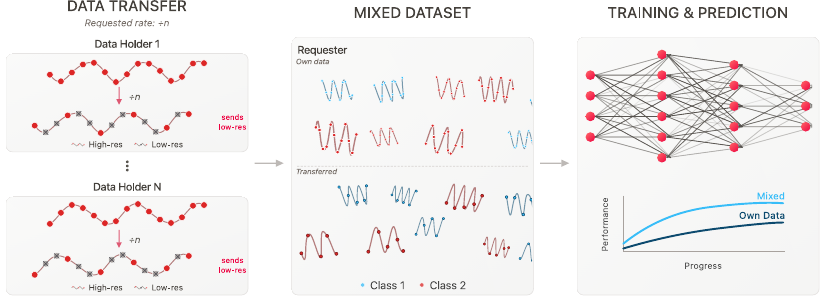}
    \caption{Illustration of the formation of a mixed-resolution dataset. A central party (the requester) possesses a limited amount of high-resolution data and requests additional data from multiple external data holders. Due to bandwidth and storage constraints, the exchanged data are provided at a lower resolution agreed upon by all parties. The resulting dataset consists of the requester’s high-resolution samples combined with low-resolution data from external sources, on which model training is subsequently performed.}
    \label{fig:system overview}
\end{figure}

A representative example arises in wearable devices, such as smart watches and hearing aids, which continuously collect data in everyday use \citep{christensen-RealWorldHearing-2021, pati-PrivacyPreservation-2024, wani-FEDEHRPrivacyPreserving-2025}. Privacy constraints, limited bandwidth, and energy resources prevent storage and transmission of raw high-resolution data. Instead, the devices only transmit feature-based representations or temporally downsampled summaries \citep{pontoppidan_data-driven_2018, zhao-ProtectingInfinite-2024}. For example, a wearable device may monitor movement or acoustic conditions continuously, but if only a coarse temporal summary is transmitted, short-lived changes and other fine-grained patterns in the original signal are no longer retained, yielding the mixed-resolution setting illustrated in Figure \ref{fig:system overview}. As a result, datasets collected from multiple sources may contain a mixture of high- and low-resolution datapoints. Despite the prevalence of such settings, the question of how much task-relevant information is retained in low-resolution data remains largely unexplored. Existing work on low-resolution learning has primarily focused on super-resolution or on improving robustness and efficiency at inference time, rather than understanding the role of low-resolution data during training under storage and privacy constraints \citep{zangeneh-LowResolution-2020, ge-LowResolutionFace-2019}.

In this work, we treat low-resolution observations not as a nuisance to be corrected, but as a deliberate design choice motivated by practical constraints. We consider a setting in which the training data comprise both high- and low-resolution datapoints. Each datapoint is assumed to be generated at a single resolution and is therefore only available in either high- or low-resolution forms. We investigate how much task-relevant information is preserved under downsampling, and to what extent low-resolution data contribute to models evaluated on high-resolution inputs. This is carried out by a theoretical analysis studying how the contribution of a datapoint would differ if it were realised at a different resolution. We validate these results empirically through simulations. Complementing this analysis, we investigate the practical usefulness of low-resolution data empirically using two model architectures: a Vision Transformer (ViT) and a Convolutional Neural Network (CNN). While both architectures can accommodate inputs of varying spatial resolution, they do so in fundamentally different ways.
ViTs accommodate changing input resolution through patch tokenisation and positional embedding (PE) interpolation, whereas CNNs require an adaptive pooling prior to classification. Comparing the two allows us to assess how architectural inductive biases interact with the information lost under downsampling.

Our main contributions are
\begin{enumerate}
    \item We provide a theoretical characterisation of the relative influence of high- and low-resolution versions of a datapoint on the model parameter distribution, and derive bounds that relate this influence to the information missing from the lower-resolution representation.
    \item Through experiments on standard benchmarks and on naturally heterogeneous medical data, we demonstrate that low-resolution data
    can improve performance on high-resolution evaluation tasks when high-resolution data is
    scarce.
\end{enumerate}

\section{Related Work}
Prior methods for learning across resolutions typically either require access to samples at multiple resolutions, perform testing purely in low-resolution regimes, or focus on robustness to resolution shifts. In contrast, we study mixed-resolution training without paired observations, focusing on the performance on high-resolution inputs at test time.

\paragraph{Super-Resolution and Cross-Resolution Learning}
Super-resolution methods aim to reconstruct high-resolution images from low-resolution inputs \citep{saharia-ImageSuperResolution-2022, zamir-LearningEnriched-2020, wang-DeepNetworks-2015, hui-LightweightImage-2019} and are sometimes used as a preprocessing step to improve downstream performance \citep{koziarski-ImpactLow-2018, wang-StudyingVery-2016}. While effective in certain regimes, such approaches are misaligned with storage- and resource-constrained settings. 

Cross-resolution learning frameworks typically rely on explicit alignment between high- and low-resolution representations. This is obtained either through the use of super-resolution pretraining in conjunction with cross-resolution feature transfer \citep{wang-StudyingVery-2016}, employing a staged fine-to-coarse training pipeline \citep{peng-FinetocoarseKnowledge-2016}, or adjusting crop-resolutions at test-time to fix the discrepancies in resolution between train and test time \citep{touvron2019fixing}, all of which requires access to the original high-resolution data examples. Such assumptions are impractical when low-resolution data originates from external sources, and upscaling low-resolution samples to fit standard CNN architectures is storage-inefficient. While \cite{koziarski-ImpactLow-2018} show that accuracy degrades with decreasing resolution, they do not consider training across multiple resolutions.

\paragraph{Low-Resolution Recognition}
Low-resolution face recognition has been addressed via  multidimensional scaling \citep{mudunuri-LowResolution-2016, yang-DiscriminativeMultidimensional-2018}, discriminative subspace alignment with multiple CNNs \citep{lu-DeepCoupled-2018, zangeneh-LowResolution-2020}, simultaneous discriminant analysis \citep{chu-LowresolutionFace-2017}, and knowledge distillation \citep{ge-LowResolutionFace-2019}. While effective under cross-resolution mismatch, these methods focus on inference in low-resolution regimes rather than exploiting low-resolution data to improve high-resolution performance.

\paragraph{Learning from Compressed Representations}
A smaller body of work considers learning directly from compressed or downsampled inputs. RandNet \citep{chang-RandnetDeep-2019} enables supervised learning from low-resolution data and connects to dictionary learning. Earlier theoretical work by \cite{fabisch-LearningCompressed-2013} shows that representing a network’s weights using a lower-dimensional parameterisation is mathematically equivalent to projecting its inputs into a lower-dimensional space. However, these approaches do not address whether training on mixed resolutions improves generalisation to high-resolution inputs.

\paragraph{Multi-Resolution Training of Transformers}
Multi-resolution training has been explored in time-series transformers \citep{chen-PathformerMultiscale-2024, wang-MedformerMultiGranularity-2024, zhang-MultiresolutionTimeSeries-2024}, where varying patch sizes capture patterns at different temporal scales. In computer vision, ResFormer \citep{tian-ResFormerScaling-2023} enforces scale consistency by replicating batches across resolutions in conjunction with a scale consistency loss, ViTAR \citep{fan-ViTARVision-2024} introduces fuzzy PEs to improve robustness to resolution shifts, and NaViT \citep{NEURIPS2023_06ea400b} uses sequence packing during training to be able to handle different input resolutions. Although these methods improve cross-resolution robustness, they do not examine the benefit of incorporating unpaired low-resolution data during training. 

\paragraph{CNNs and Vision Transformers}
CNNs \citep{lecun-BackpropagationApplied-1989} process images through convolutional layers that apply learnable local filters across spatial neighbourhoods. This hierarchical structure enables the extraction of progressively more complex features, from edges and textures in early layers to high-level semantic representations in deeper layers. Weight sharing and translation equivariance provide strong inductive biases. In contrast, ViTs \citep{dosovitskiy-ImageWorth-2021} treat images as sequences of fixed-size patches processed by Transformer encoder layers with PEs. Self-attention enables global interactions across patches from the first layer onward, although ViTs lack the locality and equivariance biases of CNNs.

\section{Problem Formulation}\label{sec:problem}
We study supervised learning in a setting where training data is not available at a single common resolution, while models are ultimately evaluated on high-resolution inputs only. Let $\mathcal{S}(\mathcal{D})=\{S(d): d\in \mathcal{D}\}$ be a stochastic process defined on a continuous domain $\mathcal{D}=[0,M]$, and let a datapoint $\boldsymbol{x}$ be a finite-dimensional representation of a realisation of the process $\mathcal{S}$. Let $s$ and $t$ denote two integer sampling rates where $s > t$. A high-resolution datapoint $\boldsymbol{x}_h \in \mathbb{R}^s$ is represented by $s$ coefficients, while a low-resolution datapoint $\tilde{\boldsymbol{x}}_l \in \mathbb{R}^t$ is represented by $t$ coefficients. For mathematical convenience and to compare the two in a common space, $\mathbb{R}^s$, we introduce an upsampling operator $f:\mathbb{R}^t \to \mathbb{R}^s$ and define $\boldsymbol{x}_l := f(\tilde{\boldsymbol{x}}_l) \in \mathbb{R}^s$. The operator preserves the alignment of the coefficients, such that, in this representation, $\boldsymbol{x}_h$ admits the decomposition: $\boldsymbol{x}_h = \boldsymbol{x}_l + \boldsymbol{x}_r$. Here $\boldsymbol{x}_r \in \mathbb{R}^s$ denotes the residual component, i.e. the information present in the high-resolution representation but absent in the low-resolution one. For simplicity, we consider processes on a one-dimensional domain, such that each datapoint is a vector. Higher-dimensional signals are handled by vectorising the coefficients.

Let $\mathcal{X}=\{\boldsymbol{x}_1, \boldsymbol{x}_2, \hdots, \boldsymbol{x}_N\}$ be a training dataset consisting of a mixture of high- and low-resolution datapoints, where each datapoint is observed at exactly one resolution. We consider a parametric model with parameters $\boldsymbol{\theta} \in \Theta$ trained using a loss function $\ell(\boldsymbol{\theta}, \boldsymbol{x})$. For notational simplicity, we write $\ell(\boldsymbol{\theta},\boldsymbol{x})$ as shorthand for $\ell(\boldsymbol{\theta}; \boldsymbol{x},y)$, i.e. the target $y$ is suppressed whenever it is fixed and unambiguous. All derivatives with respect to the input are taken with respect to the observation $\boldsymbol{x}$, not the target $y$. For a new datapoint, $\boldsymbol{x}_i$, we define its influence  as the change in the parameter distribution induced by its inclusion: 
\begin{align}
     \mathrm{KL}_i := \mathrm{KL}\big(p(\boldsymbol{\theta} \mid \mathcal{X})\,\|\,p(\boldsymbol{\theta} \mid \mathcal{X}\cup\{\boldsymbol{x}_i\})\big), 
\end{align}
where KL denotes the Kullback-Leibler divergence \citep{kullback_kullback-leibler_1951}. In the following analysis, we assume $\mathrm{KL}_i > 0$ as well as treat $\boldsymbol{x}_h$ and $\boldsymbol{x}_l$ as fixed realisations of the same underlying datapoint at different resolutions. 

\textbf{Our central question} is how the influence of a datapoint changes when it is observed at a different resolution. We characterise this difference by relating $\mathrm{KL}_h$ and $\mathrm{KL}_l$ to the residual component $\boldsymbol{x}_r$, which represents the information present in $\boldsymbol{x}_h$, but absent from $\boldsymbol{x}_l$. In contrast to prior work, we do not assume access to paired high- and low-resolution samples, nor do we require high-resolution reconstruction. Although our empirical evaluation focuses on classification, the theoretical framework permits any loss satisfying the stated regularity conditions. Dense prediction tasks, such as segmentation and localisation, require additional care because downsampling may also alter the spatial representation of the target. We therefore leave their empirical study to future work.

\section{Theoretical Analysis}
In this section, we theoretically analyse the effect of adding a high- versus a low-resolution datapoint to the training set. Following \cite{sablayrolles-WhiteboxVs-2019} and \cite{dong2022privacy} we model the parameters $\boldsymbol{\theta}$ as a random variable whose distribution depends on the training set  $\mathcal{X}$. Randomness arises from the training procedure in the form of stochastic gradients, random initialisation, and mini-batch sampling. 
\begin{assumption}[Gibbs Distribution]\label{ass:gibbs}
    Given a dataset $\mathcal{X}$ and a loss function $\ell(\boldsymbol{\theta}, \boldsymbol{x})$, the distribution of model parameters, $\boldsymbol{\theta}$, is given by 
\begin{align}
    p(\boldsymbol{\theta}|\mathcal{X}) = \frac{1}{Z(\mathcal{X})} \exp{\left(-\frac{1}{\gamma}\sum_{\boldsymbol{x}\in \mathcal{X}}\ell(\boldsymbol{\theta}, \boldsymbol{x})\right)},        
\end{align}
where the partition function $Z(\mathcal{X})=\int_{\Theta} \exp{\left(-\frac{1}{\gamma}\sum_{\boldsymbol{x}\in \mathcal{X}}\ell(\boldsymbol{\theta}, \boldsymbol{x})\right)} \, d\boldsymbol{\theta}$ is assumed to be finite. 
\end{assumption}
Without loss of generality we set the temperature parameter $\gamma =1$ for the remainder of the paper. To facilitate the analysis, we impose regularity conditions on the loss function. 

\begin{assumption}[Properties of the Loss Function]\label{ass:moments}
For a loss function $\ell(\boldsymbol{\theta}, \boldsymbol{x})$ and constants $C_1, C_2, C_3 <\infty$  
\begin{itemize}
    \item[\textbf{(A)}] \textbf{Bounded Moments} 
\begin{align*}
    &\mathbb{E}_{q(\boldsymbol{\theta})}
    \big[|\ell(\boldsymbol{\theta},\boldsymbol{x})|\big]
    \le C_1, 
    && \mathbb{E}_{q(\boldsymbol{\theta})}\big[\exp(\ell(\boldsymbol{\theta},\boldsymbol{x}))\big]
    \le C_2,
    &&
    \mathbb{E}_{q(\boldsymbol{\theta})}
    \big[\ell(\boldsymbol{\theta},\boldsymbol{x})^2\big]
    \le C_3 
    \\
    \mathrm{for} \quad  &\boldsymbol{x}\in\{\boldsymbol{x}_h,\boldsymbol{x}_l\},
    &&
    q(\boldsymbol{\theta}) \in
    \{p(\boldsymbol{\theta} \mid \mathcal{X}),
    p(\boldsymbol{\theta} \mid \mathcal{X} \cup \{\boldsymbol{x}_l\})\}.
\end{align*}
 \item[\textbf{(B)}] \textbf{Local Linearity} $\ell(\boldsymbol{\theta},\boldsymbol{x})$ is twice continuously differentiable in
$\boldsymbol{x}$ and locally linear (see Definition~\ref{itm:A1} in Appendix \ref{app:theoretical results}).
\end{itemize}

\end{assumption}
We will now state the first result. 
The ratio of KL-divergences for including a point at either high- or low-resolution admits a closed-form expression as given by the following proposition (Proof in Appendix \ref{app:ecact KL}). 
\begin{proposition}[Exact KL-divergence]\label{lem:exact KL} 
Under Assumption \ref{ass:gibbs} the ratio of KL-divergences is given by: 
\begin{align}\label{eq:kl exact}
 \frac{\mathrm{KL}_h}{\mathrm{KL}_l} &= \frac{\log{(\mathbb{E}_{{\boldsymbol{\theta} \mid \mathcal{X}}}[\exp{(-\ell(\boldsymbol{\theta}, \boldsymbol{x}_h))}])}+\mathbb{E}_{{\boldsymbol{\theta} \mid \mathcal{X}}}[\ell(\boldsymbol{\theta}, \boldsymbol{x}_h)]}{\log{(\mathbb{E}_{{\boldsymbol{\theta} \mid \mathcal{X}}}[\exp{(-\ell(\boldsymbol{\theta}, \boldsymbol{x}_l))}])}+\mathbb{E}_{{\boldsymbol{\theta} \mid \mathcal{X}}}[\ell(\boldsymbol{\theta}, \boldsymbol{x}_l)]}.
\end{align}
\end{proposition}
Proposition \ref{lem:exact KL} characterises datapoint influence through the behaviour of the loss under the posterior distribution over parameters. A ratio above $1$ means that the high-resolution point exerts a larger influence than its low-resolution counterpart, while a ratio of $1$ means there is no gain in the extra information in the high-resolution representation.  

Recognising the cumulant generating function \citep{kendall_advanced_1943} in \eqref{eq:kl exact} reveals that datapoint influence is directly affected by higher-order moments. Approximating it to second order shows that the leading effect is governed by the variance of the loss as seen in the following proposition (proof in Appendix \ref{app:var fraction}). 
\begin{proposition}[Variance Approximation]\label{cor:var fraction}
    Under assumptions \ref{ass:gibbs} and \ref{ass:moments}A, the ratio of KL-divergences admits the second-order approximation
    \begin{align}
        \frac{\mathrm{KL}_h}{\mathrm{KL}_l} \approx \frac{\mathrm{Var}_{\boldsymbol{\theta} | \mathcal{X}}[\ell(\boldsymbol{\theta}, \boldsymbol{x}_h)]}{\mathrm{Var}_{\boldsymbol{\theta} | \mathcal{X}}[\ell(\boldsymbol{\theta}, \boldsymbol{x}_l)]}. 
    \end{align}
\end{proposition}
This result provides an interpretable measure of datapoint influence. Intuitively, a datapoint must exert a larger influence on parameter distribution if the loss changes greatly across different realisations. The following proposition provides bounds of the ratio of KL-divergences, showing that the relative influence of a high-resolution datapoint is controlled by the residual component $\boldsymbol{x}_r$ (proof  is found in Appendix~\ref{app:KL ratio}).  
\begin{proposition}[Ratio of KL-divergence]\label{thm: KL ratio}
   Under assumptions \ref{ass:gibbs} and \ref{ass:moments}, the ratio of KL-divergence satisfies the approximate bounds 
    \begin{align}
     \left(1 - \sqrt{\frac{\|\boldsymbol{x}_r\|^2_{\Sigma_{\boldsymbol{g}}}}{\sigma_l^2}}\right)^2 \leq \frac{\mathrm{KL}_h}{\mathrm{KL}_l} &\leq \left(1 + \sqrt{\frac{\|\boldsymbol{x}_r\|^2_{\Sigma_{\boldsymbol{g}}}}{\sigma_l^2}}\right)^2, 
\end{align}
where $\sigma_l^2\! =\! \mathrm{Var}_{\boldsymbol{\theta} \mid \mathcal{X}}[\ell(\boldsymbol{\theta}, \boldsymbol{x}_l)]\!$ , $\!\boldsymbol{g} \!=\! \nabla_{\boldsymbol{x}}\ell(\boldsymbol{\theta}, \boldsymbol{x})\big|_{\boldsymbol{x}=\boldsymbol{x}_l}\!$ and $\Sigma_{\boldsymbol{g}}$ the covariance matrix of $\boldsymbol{g}$ under $p(\boldsymbol{\theta} \mid \mathcal{X})$. 
\end{proposition}

While Proposition~\ref{thm: KL ratio} characterises the relative influence of high- and low-resolution representations, it is also useful to quantify their absolute difference. Proposition \ref{thm: Delta KL} provides bounds on the difference between the two KL-divergences, which measures the additional contribution of the high-resolution representation (proof in Appendix~\ref{app:Delta KL}). 

\begin{proposition}[Difference in KL-divergence]\label{thm: Delta KL}
Under assumptions \ref{ass:gibbs} and \ref{ass:moments},  the difference in KL-divergences admits the approximate bounds, 
\begin{align}
        \frac{1}{2}\|\boldsymbol{x}_r\|_{\Sigma^l_{\boldsymbol{g}}}^2 - \zeta\|\boldsymbol{x}_r\|_{\Sigma^l_{\boldsymbol{g}}} \leq \mathrm{KL}_h - \mathrm{KL}_l \leq \frac{1}{2}\|\boldsymbol{x}_r\|_{\Sigma^l_{\boldsymbol{g}}}^2 + \zeta\|\boldsymbol{x}_r\|_{\Sigma^l_{\boldsymbol{g}}}, 
    \end{align}
where $\zeta = \sqrt{\mathrm{Var}_{\boldsymbol{\theta} \mid \mathcal{X} \cup \{\boldsymbol{x}_l\}}[\exp{\left(\ell(\boldsymbol{\theta}, \boldsymbol{x}_l)\right)}]} /  \mathbb{E}_{\boldsymbol{\theta} \mid \mathcal{X}\cup \{\boldsymbol{x}_l\}}[\exp{\left(\ell(\boldsymbol{\theta}, \boldsymbol{x}_l)\right)}]$, $\boldsymbol{g} = \nabla_{\boldsymbol{x}}\ell(\boldsymbol{\theta}, \boldsymbol{x})\big|_{\boldsymbol{x}=\boldsymbol{x}_l} $ and $\Sigma^l_{\boldsymbol{g}}$ is the covariance matrix of $\boldsymbol{g}$  under $p(\boldsymbol{\theta} \mid \mathcal{X}\cup \{\boldsymbol{x}_l\})$. 
\end{proposition}
Taken together, propositions~\ref{thm: KL ratio} and~\ref{thm: Delta KL} characterise both the relative and absolute benefit of including a datapoint at higher resolution. The former shows that the relative gain depends on the directional residual norm   $\|\boldsymbol{x}_r\|_{\Sigma_{\boldsymbol{g}}}$, normalised by $\sigma_l^2$. In particular, when $\sigma_l^2$ is large, the already big influence of the datapoint makes the extra coefficients in $\boldsymbol{x}_r$ less important.  Proposition~\ref{thm: Delta KL} complements this with a correction term governed by $\zeta$. Thus the value of higher resolution depends both on the discarded information and on the sensitivity of the loss to that information. 

The analysis does not impose any restriction on the resolutions present in $\mathcal{X}$, and does not depend on the particular choice of coefficients defining $\boldsymbol{x}_l$, provided alignment with $\boldsymbol{x}_h$ is preserved. $\boldsymbol{x}_l$ may therefore represent a partial feature observation with $\boldsymbol{x}_r$ being the missing features. 

The bounds in Propositions ~\ref{thm: KL ratio} and~\ref{thm: Delta KL} are obtained by controlling covariance terms via the Cauchy-Schwarz inequality \citep{gut_probability_2013}, which yields worst-case bounds. If these covariance terms are known to be non-negative, the lower bounds can be improved. The resulting tighter bounds are stated in Corollary~\ref{cor:tight bounds} with proofs in Appendix \ref{sec:tigher lower bounds}.
\begin{corollary}[Tighter Lower Bounds]\label{cor:tight bounds}
Under Assumptions~\ref{ass:gibbs} and~\ref{ass:moments}, suppose additionally that $
\mathrm{Cov}_{\boldsymbol{\theta} \mid \mathcal{X}}
\!\left[\boldsymbol{g}^\top \boldsymbol{x}_r,\,
\ell(\boldsymbol{\theta}, \boldsymbol{x}_l)\right] \ge 0$
and $\mathrm{Cov}_{\boldsymbol{\theta} \mid \mathcal{X}\cup \{\boldsymbol{x}_l\}}
\!\left[\boldsymbol{g}^\top \boldsymbol{x}_r,\,
\exp\!\left(\ell(\boldsymbol{\theta}, \boldsymbol{x}_l)\right)\right] \ge 0$. 
Then propositions~\ref{thm: KL ratio} and~\ref{thm: Delta KL} admit the tighter lower bounds
\begin{align}
   1 +\frac{\|\boldsymbol{x}_r\|_{\Sigma_{\boldsymbol{g}}}^2}{\sigma_l^2} \leq  \frac{\mathrm{KL}_h}{\mathrm{KL}_l}, \qquad
    \tfrac{1}{2}\|\boldsymbol{x}_r\|_{\Sigma^{l}_{\boldsymbol{g}}}^2 \leq \mathrm{KL}_h - \mathrm{KL}_l ,
\end{align}
where $ \sigma_l^2 \!=\! \mathrm{Var}_{\boldsymbol{\theta}\mid \mathcal{X}}[\ell(\boldsymbol{\theta}, \boldsymbol{x}_l)]$, $\boldsymbol{g}\! = \!\nabla_{\boldsymbol{x}}\ell(\boldsymbol{\theta}, \boldsymbol{x})\big|_{\boldsymbol{x}=\boldsymbol{x}_l} $,
and $\Sigma_{\boldsymbol{g}}$ and $\Sigma^l_{\boldsymbol{g}}$ are the covariance matrices of $\boldsymbol{g}$ under $p(\boldsymbol{\theta} \mid \mathcal{X})$ and $p(\boldsymbol{\theta} \mid \mathcal{X} \cup \{\boldsymbol{x}_l\})$ respectively. 
\end{corollary}
Under these non-negative covariance assumptions, the lower bound on the KL-ratio is strictly above $1$, and the lower bound on the KL-difference is strictly positive whenever $\boldsymbol{x}_r \neq \boldsymbol{0}$. Thus, in this regime, the high-resolution datapoint is more influential than its low-resolution counterpart. The proofs of Propositions~\ref{thm: KL ratio} and \ref{thm: Delta KL} rely on a first-order Taylor expansion. Appendix~\ref{app:second order} derives corresponding bounds using a second-order expansion and shows through simulations that this refinement does not produce better bounds. 

To illustrate the framework in a setting with exact KL-influence, we consider linear regression. Here the Gibbs distribution is Gaussian, and the resulting KL-influence is presented in Proposition \ref{thm: lin reg} with the associated proof in Appendix \ref{sec:lin reg proof}. 
\begin{proposition}[Closed-form KL-influence for Linear Regression]\label{thm: lin reg}
For a linear regression model trained using the mean squared error loss function the model parameters, under Assumption \ref{ass:gibbs}, follow a normal distribution, $\boldsymbol{\theta} \sim \mathcal{N}(\boldsymbol{\mu}, \boldsymbol{\Sigma})$. For a datapoint $(\boldsymbol{x}_i, y_i)$, the KL-influence admits the exact decomposition: 
\begin{align}
    \mathrm{KL}_i =
    & -\frac{1}{2}\log(1+2s_i^2) - \frac{(y_i - m_i)^2}{2s_i^2+1} + (y_i-m_i)^2 + s_i^2, 
\end{align}
where $m_i = \boldsymbol{x}_i^\top \boldsymbol{\mu}$ and $s_i^2 = \boldsymbol{x}_i^\top \boldsymbol{\Sigma}\boldsymbol{x}_i$. 
\end{proposition}
The proposition identifies two factors controlling datapoint influence: the squared prediction residual $(y_i - m_i)^2$ and the posterior predictive uncertainty $s_i^2$. As $s_i^2 \to 0$, then $\text{KL}_i \to 0$, whereas $\mathrm{KL}_i \to \infty$ as $s_i^2\to \infty$. Thus a datapoint is generally more influential when the posterior is more uncertain about its prediction. This is consistent with the variance-based interpretation from Proposition \ref{cor:var fraction}, while providing an exact result that does not rely on a second-order truncation. 
For linear regression the first covariance condition of Corollary \ref{cor:tight bounds} can be analysed in closed form. It is violated when $\boldsymbol{x}_r$ acts sufficiently correctively of the low-resolution prediction or uncertainty. More details can be found in Appendix~\ref{app:lin reg cov ass}. 

\subsection{Simulations of Theoretical Results}
To validate the theoretical results, we conduct a series of simulations. For this, we need a dataset where a high-resolution datapoint fulfils the decomposition $\boldsymbol{x}_h = \boldsymbol{x}_l + \boldsymbol{x}_r$. To construct such a dataset, the first 100 images from each of the first two classes of CIFAR-10 are extracted. For each of the datapoints a scalogram using the Discrete Wavelet Transform (DWT) is computed with three levels using the \texttt{db2} mother wavelet \citep{mallat_wavelet_2009}. To create different degrees of downsampled data, the bands in the scalogram containing the highest frequencies are progressively set to zero. The final datapoint, $\boldsymbol{x}_l$, is restored using the inverse DWT and projected to a $10$-dimensional vector using PCA \citep{pearson1901liii}. For more details on how this ensures that $\boldsymbol{x}_h = \boldsymbol{x}_l + \boldsymbol{x}_r$ we refer to Appendix~\ref{app:synthetic data wavelet}. 

\begin{figure}
    \centering
    \begin{subfigure}[b]{0.49\textwidth}
        \includegraphics[trim=0 0.2cm 0 0.2cm, clip=true]{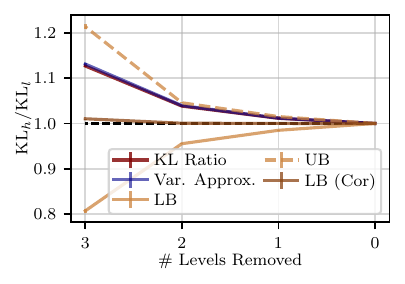}
    \end{subfigure}
    \begin{subfigure}[b]{0.49\textwidth}
        \includegraphics[trim=0 0.2cm 0 0.2cm, clip=true]{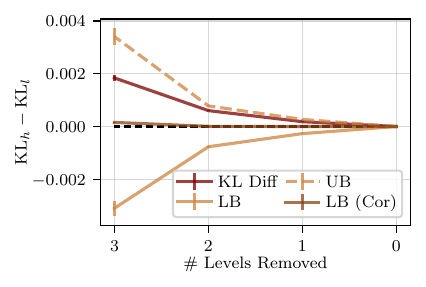}
        
    \end{subfigure}
    \caption{Simulation of bounds as a function of progressively removed high-frequency content. Error bars denote $\pm 5$ times the standard deviation. As more high-frequency components are removed, the influence of low-resolution datapoints decreases, validating the theoretical predictions. UB = upper bound, LB = lower bound. Cor denotes bounds from Corollary \ref{cor:tight bounds}. \textbf{Left}: True KL-divergence ratio, variance-based approximation from Proposition \ref{cor:var fraction}, the bounds from Proposition \ref{thm: KL ratio} and the tighter ratio lower bound from Corollary \ref{cor:tight bounds}. 
    \textbf{Right}: True KL-divergence difference, the bounds from Proposition \ref{thm: Delta KL}, and the tighter difference lower bound from Corollary \ref{cor:tight bounds}.}
    \label{fig:simulation}
\end{figure}

A neural network consisting of four linear layers with ReLU activations is then trained to distinguish between the two classes. We compute the true KL-divergences using the expression in Proposition \ref{lem:exact KL}, assess the approximation in Proposition \ref{cor:var fraction}, and compute the two bounds from propositions \ref{thm: KL ratio} and \ref{thm: Delta KL} as well as the tighter bounds in Corollary \ref{cor:tight bounds}. Results are depicted in Figure \ref{fig:simulation} and generated by training $10,000$ models repeated over five random seeds. The black dashed line in the left panel denotes a fraction of $1$, and in the right panel it denotes a difference of $0$. The $x$-axis denotes the number of levels/frequency bands that have been removed  where 0 is no removal. Exact KL-divergence values and their corresponding bounds can be found in Appendix \ref{app:simulation bounds}. 

The variance approximation captures the overall behaviour of the KL-ratio. For both the KL-ratio and KL-difference, the lower and upper bounds become looser as the magnitude of the residual component increases. The tighter bounds introduced in Corollary \ref{cor:tight bounds} are indeed tighter which is consistent with the corresponding covariance terms being non-negative. Appendix \ref{app:tightness of bounds} shows that the variance approximation remains accurate across input dimensions and model sizes with relative error below $3\%$ that decreases as model size grows. The appendix also shows how the bounds tighten with model size. Appendix \ref{app:toy example} further shows that high-resolution datapoints lying outside the support of the low-resolution distribution induce the largest parameter changes, illustrating how information discarded by downsampling governs datapoint influence.  

\section{Experimental Setup}\label{sec:experiments}
We empirically evaluate whether adding low-resolution training data leads to measurable gains in practice. We do this in both controlled and naturally heterogeneous settings focusing on regimes where high-resolution training data is scarce. Models are evaluated exclusively on high-resolution inputs, and low-resolution samples are processed at the low resolution rather than being upsampled. The low-resolution inputs thus correspond to $\tilde{x}_l\in\mathbb{R}^t$ in Section~\ref{sec:problem} avoiding the storage and computational cost of embedding them in $\mathbb{R}^s$. Unless otherwise stated, results are averaged over five independent runs. Additional implementation and training details are provided in Appendix~\ref{app:training details}. 

\paragraph{Controlled Setting}
We first conduct experiments on CIFAR10 and CIFAR100 \citep{krizhevsky-LearningMultiple-2009} to isolate the effect downsampling has on performance. This is done by both varying the proportion of high-resolution sampled as well as the resolution of the downsampled samples. This gives rise to four different types of experiments: training on a share of the high-resolution data (\textbf{Subset}); training on the same share but adding the remaining data in a low-resolution format (\textbf{Ratio}); training on a fully downsampled dataset (\textbf{Downsampled}); replacing $10\%$ of the downsampled data with full-resolution samples (\textbf{Size}). CIFAR images have sidelength of $L=32$. We use $L \in \{4, 8, 12, 16, 20\}$ in \textbf{Size} experiments and $L=8$ in \textbf{Ratio} experiments. We evaluate a ViT and a modified ResNet18 with details given in Appendix \ref{app:training details}. Results on the one-dimensional AudioMNIST dataset are reported in Appendix \ref{app:res audiomnist}. 

\paragraph{Heterogeneous Setting}
We model the setting from Figure \ref{fig:system overview} using the skin-lesion dataset Fed-ISIC-2019 \citep{tschandl2018ham10000,codella2018skin,combalia2019bcn20000,ogier2022flamby}. As the data is collected across six medical centres with substantial centre and label heterogeneity, this setting combines a resolution shift with a distribution shift. We select centre 1 as the requester and evaluate exclusively on its high-resolution test data. The requester then retains $\{10, 25, 50, 100\}\%$ of its local training data at $224 \times 224$ resolution while progressively receiving images from the remaining centres at a $56 \times 56$ resolution which we denote the \textbf{Grow} experiment. Donated samples are added beginning with minority classes. We report balanced accuracy due to the strong class imbalance. Remaining details on the model architectures and optimisation strategies are reported in Appendix \ref{app:training details}.

\section{Empirical Results}
\begin{figure}
    \centering
    \includegraphics[trim=0 0.2cm 0 0.23cm, clip=true]{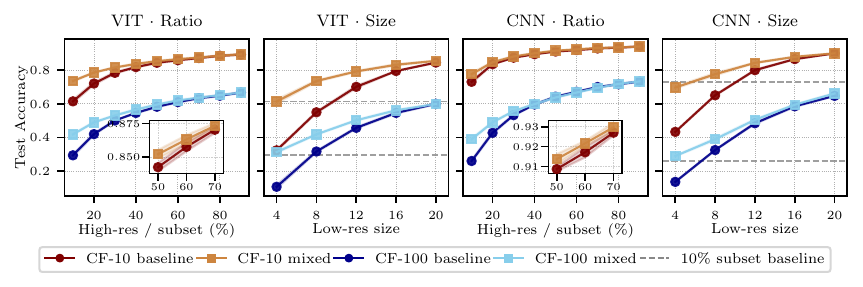}
    \caption{Effect of mixed-resolution training under different configurations. \textbf{Subset} is baseline for \textbf{Ratio}; both train on the same proportion of high-resolution samples and \textbf{Ratio} additionally includes the remaining samples in low-resolution format ($L=8$). \textbf{Downsampled} is the baseline for \textbf{Size}, comparing a fully downsampled dataset against a mixed-resolution setting with $10\%$ high-resolution and $90\%$ low-resolution samples of varying sizes. The $10\%$ subset baseline is shown in grey. CF denotes CIFAR and "Down." is short for Downsampled. Mixed-resolution training improves performance across datasets and architectures.}
    \label{fig:cifar results}
\end{figure}

\paragraph{Controlled Setting} Figure \ref{fig:cifar results} depicts the results from the CIFAR datasets. The \textbf{Ratio} experiments are compared against \textbf{Subset}, while the \textbf{Size} experiments are compared against \textbf{Downsampled}. Across architectures and datasets, mixed-resolution training consistently improves performance, with gains diminishing as the proportion of high-resolution data increases. The ViT seems to benefit more for these datasets from the addition of low-resolution samples. One possible explanation is that CNNs rely more heavily on local, spatial structure, which can be disrupted by coarse downsampling. This indicates that low-resolution data can help boost performance when it originates from the same distribution. Appendix \ref{app:ablation studies} contains different ablation studies including one that demonstrates that low-resolution data does provide a true learning signal, rather than just acting as a regulariser. 
\begin{figure}
    \centering
    \includegraphics[trim=0 0.3cm 0 0.22cm, clip=true]{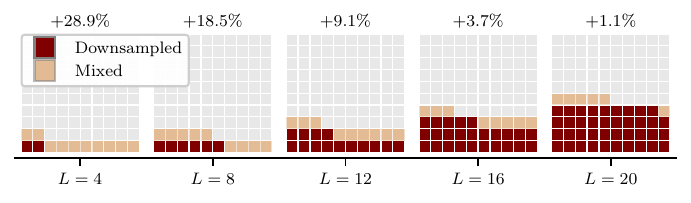}
    \caption{Trade-off between accuracy gain and storage cost for the \textbf{Size} experiment compared to the \textbf{Downsampled} on CIFAR-10 using the ViT. The $10\times 10$ grid represents storage of the full dataset at full resolution, red cells correspond to storage of the \textbf{Downsampled} regime, and yellow cells to additional storage for the mixed-resolution (\textbf{Size}) regime.  For $L=12$, the downsampled dataset uses $12^2/32^2=0.14$ of full storage, while the mixed dataset uses $(12^2\cdot0.9+32^2\cdot0.1)/32^2=0.226$. Storage ratios are rounded to the nearest percentage point. The percentage above each panel denotes the relative improvement in classification accuracy achieved by the mixed approach. }
    \label{fig:storage size experiment}
\end{figure}
While Figure~\ref{fig:cifar results} highlights the performance benefits of incorporating low-resolution data, it does not account for the associated storage costs. 
Figure~\ref{fig:storage size experiment} addresses this trade-off by visualising the additional storage required by the \textbf{Size} experiment compared with the \textbf{Downsampled} setting.  The figure shows that even small amounts of high-resolution data can yield substantial improvements in accuracy.

Taken together, figures~\ref{fig:cifar results} and~\ref{fig:storage size experiment} illustrate the performance-storage trade-offs of mixed-resolution training. The \textbf{Ratio} experiment shows that, when high-resolution data is scarce, collecting more low-resolution data can be more beneficial than acquiring a few additional high-resolution samples. Conversely, the \textbf{Size} experiment shows that selectively adding high-resolution samples can be more effective than uniformly increasing the resolution of the entire dataset. 

\begin{figure}
    \centering    \includegraphics{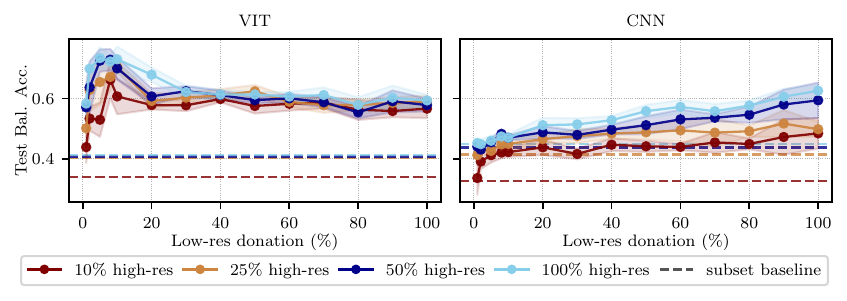}
    \caption{\textbf{Subset} and \textbf{Grow} experiments on the Fed-ISIC-2019 dataset. Each curve represents a different amount of high-resolution data from the requester, and the horizontal axis denotes the amount of low-resolution data that has been donated. Dotted lines denote corresponding high-resolution-only \textbf{Subset} baselines. }
    \label{fig:fedisic results}
\end{figure}

\paragraph{Heterogeneous Setting} Figure \ref{fig:fedisic results} compares high-resolution baselines with progressively adding low-resolution data from external sources. The CNN generally benefits from additional donations, particularly when requester data is scarce, showing that low-resolution data can remain useful under centre and label heterogeneity. For the ViT, small donations improve performance, but the gains saturate or reverse as more is added. To disentangle the roles of resolution and distribution shift, Appendix \ref{app:res audiomnist} repeats the experiment with donor data shared at full resolution, using the same donation order. In this setting, the decline largely disappears for larger requester subsets, and is weaker for the $10\%$ setting. Together with the controlled experiments in Figure \ref{fig:cifar results}, where the ViT benefits from low-resolution data from the same distribution, this suggests that neither resolution- nor distribution shift alone accounts for the degradation, but that their combination does. The CNN appears more robust to this combination. 
Appendix \ref{app:ablation studies} shows that these gains persist when donor data is additionally corrupted by blur or noise. 

\section{Conclusion}
In this work, we studied mixed-resolution training, where only a subset of the training data is available at high resolution, while the remaining data is observed in a downsampled form and evaluation is performed on high-resolution inputs. We showed theoretically that the contribution of a high-resolution datapoint relative to its low-resolution counterpart is governed both by the information missing from the low-resolution representation and by the sensitivity of the loss to that missing information. 
Empirically, low-resolution data consistently improves performance in controlled experiments and remains useful under the natural centre and label heterogeneity of Fed-ISIC-2019.
Overall our results demonstrate low-resolution data as a useful learning resource. 
\section*{Limitations}\label{sec:limitations}
Our bounds rely on a local linear approximation that is expected to be most accurate when high- and low-resolution representations remain sufficiently close. Our experiments only consider classification. While the theory is agnostic to the choice of loss function provided it satisfies the regularity conditions, segmentation and localisation tasks require additional consideration because downsampling may alter the spatial representation of the target.

\bibliographystyle{plainnat} 
\bibliography{references}

\appendix
\section*{Appendix}
\section{Theoretical Results}\label{app:theoretical results}
We first introduce the definitions and identities used in the proofs.
\subsection{Preliminary}\label{app:preliminary}

Additional definitions 
\begin{enumerate}[label=\textbf{(A\arabic*)}, ref=\textbf{A\arabic*}]
    \item \label{itm:A0}
    The Kullback-Leibler divergence for two probability distributions $p$ and $q$ has the following definition \citep{kullback_kullback-leibler_1951}: 
    \begin{align}
        \mathrm{KL}(p \mid\mid q) = \int p(\boldsymbol{\theta}) \log\left(\frac{p(\boldsymbol{\theta})}{q(\boldsymbol{\theta})}\right)\, d \boldsymbol{\theta}. 
    \end{align}
    \item \label{itm:A1}
    A function, $f$, is locally linear if it can be written as: 
    \begin{align}
        f(\boldsymbol{x}) = f(\boldsymbol{a}) + \nabla^\top f(\boldsymbol{a}) (\boldsymbol{x} - \boldsymbol{a}) + h_1(\boldsymbol{x})\,\|\boldsymbol{x} - \boldsymbol{a}\|, 
        \quad \lim_{\boldsymbol{x} \to \boldsymbol{a}} h_1(\boldsymbol{x}) = 0.
    \end{align}
    In the following derivations we consider the $h_1$ term to be negligible. 
    \item \label{itm:A2}
    The Cauchy–Schwarz inequality for variances for random variables $X$ and $Y$ is given by \citep{gut_probability_2013}: 
    \begin{align}
        |\mathrm{Cov}[X, Y]| &\leq \sqrt{\mathrm{Var}[X]\mathrm{Var}[Y]}\iff \\ -\sqrt{\mathrm{Var}[X]\mathrm{Var}[Y]} &\leq \mathrm{Cov}[X, Y] \leq \sqrt{\mathrm{Var}[X]\mathrm{Var}[Y]}
    \end{align}

    \item \label{itm:A3}
    The variance of a vector product for a vector $\boldsymbol{a}$ and $\boldsymbol{x}$ is $\mathrm{Var}[\boldsymbol{a}^\top \boldsymbol{x}] = \boldsymbol{x}^\top \Sigma_{\boldsymbol{a}}\boldsymbol{x}$, where $\Sigma_{\boldsymbol{a}}$ is the covariance matrix of $\boldsymbol{a}$:
    \begin{align*}
    \mathrm{Var}[\boldsymbol{a}^\top \boldsymbol{x}]
    &=\mathbb{E}\left[\left(\boldsymbol{a}^\top \boldsymbol{x} - \mathbb{E}[\boldsymbol{a}^\top \boldsymbol{x}]\right)^2\right] \\ 
    &= \mathbb{E}\left[\left(\boldsymbol{a}^\top \boldsymbol{x} - \mathbb{E}[\boldsymbol{a}]^\top \boldsymbol{x}\right)^2\right] \\ 
    &=\mathbb{E}\left[\left((\boldsymbol{a}^\top - \mathbb{E}[\boldsymbol{a}]^\top)\boldsymbol{x}\right)^2\right] \\ 
    &= \mathbb{E}\left[\boldsymbol{x}^\top (\boldsymbol{a}^\top - \mathbb{E}[\boldsymbol{a}]^\top )^\top (\boldsymbol{a}^\top - \mathbb{E}[\boldsymbol{a}]^\top ) \boldsymbol{x}\right] \\ 
    &= \mathbb{E}\left[\boldsymbol{x}^\top (\boldsymbol{a} - \mathbb{E}[\boldsymbol{a}]) (\boldsymbol{a}- \mathbb{E}[\boldsymbol{a}])^\top \boldsymbol{x}\right] \\ 
    &=\boldsymbol{x}^\top\mathbb{E}\left[ (\boldsymbol{a} - \mathbb{E}[\boldsymbol{a}]) (\boldsymbol{a}- \mathbb{E}[\boldsymbol{a}])^\top \right] \boldsymbol{x} \\
    &= \boldsymbol{x}^\top \Sigma_{\boldsymbol{a}}\boldsymbol{x}
    \end{align*}

    \item \label{itm:A4}
    Let $\|\boldsymbol{x}\|_A$ denote the Mahalanobis distance from the zero vector to a point $\boldsymbol{x}$ given by $\boldsymbol{x}^\top A\boldsymbol{x} = \|\boldsymbol{x}\|_{A}^2$ for a positive semi-definite matrix $A$ \citep{galeano_mahalanobis_2015}. 
\end{enumerate}

A datapoint consists of an observation $\boldsymbol{x}$ and a target $y$. The loss function is written in short as $\ell(\boldsymbol{\theta}, \boldsymbol{x}) = \ell(\boldsymbol{\theta}; \boldsymbol{x}, y)$ such that the target is implicit. For all derivatives it is implied they are taken with respect to the observation, $\boldsymbol{x}$ and not the target. 

Throughout the theoretical analysis, the datapoints $\boldsymbol{x}_h$ and $\boldsymbol{x}_l$ are treated as fixed, and the only source of randomness are the model parameters $\boldsymbol{\theta}$. 


\subsection{Proof of Proposition \ref{lem:exact KL}} \label{app:ecact KL}
Let $\mathcal{X} = \{\boldsymbol{x}_1, \boldsymbol{x}_2, \hdots, \boldsymbol{x}_N\}$ be a dataset consisting of samples of different resolutions. We introduce $\mathcal{X}_h = \mathcal{X} \cup \{\boldsymbol{x}_h\}$ and $\mathcal{X}_l = \mathcal{X} \cup \{\boldsymbol{x}_l\}$, where $\boldsymbol{x}_h$ and $\boldsymbol{x}_l$ denote the same underlying datapoint observed at high and low resolution, respectively.

Under Assumption \ref{ass:gibbs}, the model parameters conditioned on the dataset $\mathcal{X}$ follow the distribution: 
\begin{align}
    p(\boldsymbol{\theta} \mid \mathcal{X}) = \frac{1}{Z(\mathcal{X})} \exp{\left(-\frac{1}{\gamma}\sum_{i=1}^N \ell (\boldsymbol{\theta}, \boldsymbol{x}_i)\right)},  
\end{align}
where, without loss of generality, we set $\gamma=1$. 
For notational convenience, define $Z:=Z(\mathcal{X})$, $Z_h:=Z(\mathcal{X}\cup\{\boldsymbol{x}_h\})$ and $Z_l:=Z(\mathcal{X}\cup\{\boldsymbol{x}_l\})$. Then the distribution of the model parameters when adding $\boldsymbol{x}_h$ is, 
\begin{align}\label{eq:klh}
    p(\boldsymbol{\theta} \mid \mathcal{X}_h) &= \frac{1}{Z_h} \exp{\left(-\sum_{i=1}^{N} \ell(\boldsymbol{\theta}, \boldsymbol{x}_i) - \ell(\boldsymbol{\theta}, \boldsymbol{x}_h)\right)},
\end{align}
The distribution can likewise be found for adding $\boldsymbol{x}_l$. It follows that the ratio between the two distributions can be written as 
\begin{align}\label{eq:frac dists}
    \frac{p(\boldsymbol{\theta} \mid \mathcal{X})}{p(\boldsymbol{\theta} \mid \mathcal{X}_h)} = \frac{Z_h}{Z} \exp{\left(\ell(\boldsymbol{\theta}, \boldsymbol{x}_h)\right)}. 
\end{align}
The KL-divergence corresponding to the inclusion of $\boldsymbol{x}_h$
\begin{align}
    \mathrm{KL}_h = \mathbb{E}_{\boldsymbol{\theta} \mid \mathcal{X}}\left[\log\left(\frac{p(\boldsymbol{\theta}\mid \mathcal{X})}{p(\boldsymbol{\theta} \mid \mathcal{X}_h)}\right)\right], 
\end{align}
which simplifies to
\begin{align}
    \mathrm{KL}_h = \log\left(\frac{Z_h}{Z}\right) + \mathbb{E}_{\boldsymbol{\theta} \mid \mathcal{X}}[\ell(\boldsymbol{\theta}, \boldsymbol{x}_h)]. 
\end{align}
An identical argument yields
\begin{align}
    \mathrm{KL}_l = \log\left(\frac{Z_l}{Z}\right) + \mathbb{E}_{\boldsymbol{\theta} \mid \mathcal{X}}[\ell(\boldsymbol{\theta}, \boldsymbol{x}_l)]. 
\end{align}
To identify the ratio $Z_h/Z$, similarly $Z_l/Z$, we note that the ratio between the two distributions can be written as \eqref{eq:frac dists}. Rearranging and integrating over both sides yields 
\begin{align}
    \int p(\boldsymbol{\theta} \mid \mathcal{X}_h)\, d\boldsymbol{\theta} &= \frac{Z}{Z_h} \int p(\boldsymbol{\theta} \mid \mathcal{X}) \exp{(-\ell(\boldsymbol{\theta}, \boldsymbol{x}_h))} \, d \boldsymbol{\theta}, 
\end{align}
hence 
\begin{align}
    \frac{Z_h}{Z} = \mathbb{E}_{\boldsymbol{\theta} \mid \mathcal{X}}[\exp{(-\ell(\boldsymbol{\theta}, \boldsymbol{x}_h))}]. 
\end{align}
Substituting these expressions into the formulas for $\mathrm{KL}_h$ and $\mathrm{KL}_l$ yields the desired result: 
\begin{align}\label{eq:cgf step}
    \frac{\mathrm{KL}_h}{\mathrm{KL}_l} &= \frac{\log{(\mathbb{E}_{{\boldsymbol{\theta} \mid \mathcal{X}}}[\exp{(-\ell(\boldsymbol{\theta}, \boldsymbol{x}_h))}])}+\mathbb{E}_{{\boldsymbol{\theta} \mid \mathcal{X}}}[\ell(\boldsymbol{\theta}, \boldsymbol{x}_h)]}{\log{(\mathbb{E}_{{\boldsymbol{\theta} \mid \mathcal{X}}}[\exp{(-\ell(\boldsymbol{\theta}, \boldsymbol{x}_l))}])}+\mathbb{E}_{{\boldsymbol{\theta} \mid \mathcal{X}}}[\ell(\boldsymbol{\theta}, \boldsymbol{x}_l)]}. 
\end{align}

\subsection{Proof of Proposition \ref{cor:var fraction}}\label{app:var fraction}
The cumulant generating function of a random variable $Y$ is given by \citep{kendall_advanced_1943}
\begin{align}
    K_Y(b)
    :=
    \log \mathbb{E}[e^{bY}]
    =
    \sum_{n=1}^{\infty}\kappa_n \frac{b^n}{n!},
\end{align}
where $\kappa_n$ denotes the $n$th cumulant of $Y$. Truncating this expansion at second order yields the approximation
\begin{align}\label{eq:truncated cgf}
    K_Y(b)
    \approx
    \kappa_1 b + \kappa_2 \frac{b^2}{2} = \mathbb{E}[Y]b + \frac{b^2}{2} \mathrm{Var}[Y].
\end{align}
For a fixed datapoint $\boldsymbol{x}$, define the random variable
\begin{align}
    Y_{\boldsymbol{x}} := \ell(\boldsymbol{\theta},\boldsymbol{x}),
    \qquad
    \boldsymbol{\theta} \sim p(\boldsymbol{\theta}\mid\mathcal X).
\end{align}
Since the expectation of a function, $h$, is $\mathbb{E}[h(x)] = \int h(x)p(x)\, dx$ then 
\begin{align}
    \mathbb{E}_{\boldsymbol{\theta}\mid\mathcal X}[h(\boldsymbol{\theta})]
    =
    \int_{\Theta} h(\boldsymbol{\theta})\, p(\boldsymbol{\theta}\mid\mathcal X)\, d\boldsymbol{\theta},
\end{align}
for any integrable function $h$. Setting $b=-1$ and $Y=Y_{\boldsymbol{x}}$ in \eqref{eq:truncated cgf} yields 
\begin{align}
    K_{Y_x}(-1)
    \approx -\mathbb{E}_{\boldsymbol{\theta} \mid \mathcal{X}}[\ell(\boldsymbol{\theta}, \boldsymbol{x})] + \frac{1}{2}\mathrm{Var}_{\boldsymbol{\theta} \mid \mathcal{X}}[\ell(\boldsymbol{\theta}, \boldsymbol{x})], 
\end{align}
since the first two cumulants satisfy $\kappa_1=\mathbb{E}[Y]$ and $\kappa_2=\mathrm{Var}[Y]$. 
Applying this to $Y_{\boldsymbol{x}_l}=\ell(\boldsymbol{\theta},\boldsymbol{x}_l)$ and $Y_{\boldsymbol{x}_h}=\ell(\boldsymbol{\theta},\boldsymbol{x}_h)$ results in 
\begin{align}
    \log \mathbb{E}_{\boldsymbol{\theta}\mid\mathcal X}
    \left[e^{-\ell(\boldsymbol{\theta},\boldsymbol{x}_l)}\right]
    &\approx
   - \mathbb{E}_{\boldsymbol{\theta}\mid\mathcal X}
    \left[\ell(\boldsymbol{\theta},\boldsymbol{x}_l)\right]
    +
    \frac{1}{2}
    \mathrm{Var}_{\boldsymbol{\theta}\mid\mathcal X}
    \left[\ell(\boldsymbol{\theta},\boldsymbol{x}_l)\right],\\
    \log \mathbb{E}_{\boldsymbol{\theta}\mid\mathcal X}
    \left[e^{-\ell(\boldsymbol{\theta},\boldsymbol{x}_h)}\right]
    &\approx
    -\mathbb{E}_{\boldsymbol{\theta}\mid\mathcal X}
    \left[\ell(\boldsymbol{\theta},\boldsymbol{x}_h)\right]
    +
    \frac{1}{2}
    \mathrm{Var}_{\boldsymbol{\theta}\mid\mathcal X}
    \left[\ell(\boldsymbol{\theta},\boldsymbol{x}_h)\right].
\end{align}
Substituting these approximations into \eqref{eq:cgf step} yields
\begin{align}\label{eq:var frac app}
     \frac{\mathrm{KL}_h}{\mathrm{KL}_l}
     &\approx
     \frac{\mathrm{Var}_{\boldsymbol{\theta} \mid \mathcal{X}}
     \left[\ell(\boldsymbol{\theta}, \boldsymbol{x}_h)\right]}
     {\mathrm{Var}_{\boldsymbol{\theta} \mid \mathcal{X}}
     \left[\ell(\boldsymbol{\theta}, \boldsymbol{x}_l)\right]},
\end{align}
which is what we wanted to show. 
In the case of $\gamma \ne 1$, \eqref{eq:kl exact} becomes
\begin{align}
    \frac{\mathrm{KL}_h}{\mathrm{KL}_l} = \frac{\log \mathbb{E}_{\theta \mid \mathcal{X}}[\exp(-\frac{1}{\gamma}\ell(\theta, x_h))] + \frac{1}{\gamma}\mathbb{E}_{\theta \mid \mathcal{X}}[\ell(\theta, x_h)]}{\log \mathbb{E}_{\theta \mid \mathcal{X}}[\exp(-\frac{1}{\gamma}\ell(\theta, x_l))] + \frac{1}{\gamma}\mathbb{E}_{\theta \mid \mathcal{X}}[\ell(\theta, x_l)]}, 
\end{align}
in which case we can choose $b=-1/\gamma$ to obtain \eqref{eq:var frac app} again. 

\subsection{Proof of Proposition \ref{thm: KL ratio}}\label{app:KL ratio}
Starting from the result from Proposition \ref{cor:var fraction}: 
\begin{align}\label{eq:cor3 starting point1}
    \frac{\mathrm{KL}_h}{\mathrm{KL}_l} &\approx \frac{\mathrm{Var}_{\boldsymbol{\theta} \mid \mathcal{X}}\left[\ell(\boldsymbol{\theta}, \boldsymbol{x}_h)\right]}{\mathrm{Var}_{\boldsymbol{\theta} \mid \mathcal{X}}\left[\ell(\boldsymbol{\theta}, \boldsymbol{x}_l)\right]}, 
\end{align}
we use (\ref{itm:A1}) on the loss function:  
\begin{align}
     \ell(\boldsymbol{\theta}, \boldsymbol{x}_h) = \ell(\boldsymbol{\theta}, \boldsymbol{x}_l + \boldsymbol{x}_r) &\approx \ell(\boldsymbol{\theta}, \boldsymbol{x}_l) + \nabla^\top_{\boldsymbol{x}}\ell(\boldsymbol{\theta}, \boldsymbol{x})\big|_{\boldsymbol{x}=\boldsymbol{x}_l}\boldsymbol{x}_r, 
\end{align}
where we introduce $\boldsymbol{g} = \nabla_{\boldsymbol{x}}\ell(\boldsymbol{\theta}, \boldsymbol{x})\big|_{\boldsymbol{x}=\boldsymbol{x}_l}$. A rewrite of the numerator of \eqref{eq:cor3 starting point1} yields, 
\begin{align}\label{eq:p5 low bound}
    \mathrm{Var}[\ell(\boldsymbol{\theta}, \boldsymbol{x}_h)] = \mathrm{Var}[\ell(\boldsymbol{\theta}, \boldsymbol{x}_l)] + \mathrm{Var}[\boldsymbol{g}^\top \boldsymbol{x}_r] + 2 \mathrm{Cov}[\ell(\boldsymbol{\theta}, \boldsymbol{x}_l), \boldsymbol{g}^\top \boldsymbol{x}_r]. 
\end{align}
Lower bound the covariance using (\ref{itm:A2}):  
\begin{align}
     \mathrm{Var}[\ell(\boldsymbol{\theta}, \boldsymbol{x}_h)]  \geq \mathrm{Var}[\ell(\boldsymbol{\theta}, \boldsymbol{x}_l)] + \mathrm{Var}[\boldsymbol{g}^\top \boldsymbol{x}_r] - 2\sqrt{\mathrm{Var}[\ell(\boldsymbol{\theta}, \boldsymbol{x}_l)] \mathrm{Var}[\boldsymbol{g}^\top \boldsymbol{x}_r]}. 
\end{align}
Introduce $\sigma_l^2:=\mathrm{Var}[\ell(\boldsymbol{\theta}, \boldsymbol{x}_l)]$ and using (\ref{itm:A3}): 
\begin{align}
     \frac{\mathrm{KL}_h}{\mathrm{KL}_l} &\geq \frac{\sigma_l^2 + \boldsymbol{x}_r^\top\Sigma_{\boldsymbol{g}}\boldsymbol{x}_r -2 \sqrt{\sigma_l^2 \boldsymbol{x}_r^\top \Sigma_{\boldsymbol{g}}\boldsymbol{x}_r}}{\sigma_l^2},
\end{align}
where using (\ref{itm:A4}) gives us the following lower bound: 
\begin{align}
    \frac{\mathrm{KL}_h}{\mathrm{KL}_l} \geq \left(1 - \sqrt{\frac{\|\boldsymbol{x}_r\|_{\Sigma_{\boldsymbol{g}}}^2}{\sigma_l^2}}\right)^2. 
\end{align}
Using (\ref{itm:A2}) to upper bound eq. \eqref{eq:p5 low bound} yields the corresponding upper bound: 
\begin{align}
    \frac{\mathrm{KL}_h}{\mathrm{KL}_l} \leq \left(1 + \sqrt{\frac{\|\boldsymbol{x}_r\|_{\Sigma_{\boldsymbol{g}}}^2}{\sigma_l^2}}\right)^2.
\end{align}

\subsection{Proof of Proposition \ref{thm: Delta KL}}\label{app:Delta KL}
Starting from Definition (\ref{itm:A0}), we can write the difference between $\mathrm{KL}_h$ and $\mathrm{KL}_l$ as
\begin{align}
    \mathrm{KL}_h -\mathrm{KL}_l &= \int p(\boldsymbol{\theta} \mid \mathcal{X})\log{\left(\frac{p(\boldsymbol{\theta} \mid \mathcal{X})}{p(\boldsymbol{\theta} \mid \mathcal{X}_h)}\right)} \, d\boldsymbol{\theta} - \int p(\boldsymbol{\theta} \mid \mathcal{X})\log{\left(\frac{p(\boldsymbol{\theta} \mid \mathcal{X})}{p(\boldsymbol{\theta} \mid  \mathcal{X}_l)}\right)} \, d\boldsymbol{\theta} \\
    &= -\int p(\boldsymbol{\theta} \mid \mathcal{X})\log \left(p(\boldsymbol{\theta} \mid  \mathcal{X}_h)\right)\, d\boldsymbol{\theta} + \int p(\boldsymbol{\theta} \mid \mathcal{X}) \log\left(p(\boldsymbol{\theta} \mid  \mathcal{X}_l)\right)\, d\boldsymbol{\theta}
    \\&= \int p(\boldsymbol{\theta} \mid \mathcal{X}) \log{\left(\frac{p(\boldsymbol{\theta} \mid  \mathcal{X}_l)}{p(\boldsymbol{\theta}\mid  \mathcal{X}_h)}\right)}\, d\boldsymbol{\theta}. \label{eq:diff eq1}
\end{align}
Under Assumption \ref{ass:gibbs}, the log term is expressed as: 
\begin{align}
    \log{\left(\frac{P(\boldsymbol{\theta} \mid \mathcal{X}_l)}{P(\boldsymbol{\theta}\mid  \mathcal{X}_h)}\right)} &= \log{\left(\frac{\frac{1}{Z_l} \exp{\left(-\sum_{i=1}^{N} \ell(\boldsymbol{\theta}, \boldsymbol{x}_i) - \ell(\boldsymbol{\theta}, \boldsymbol{x}_l)\right)}}{\frac{1}{Z_h} \exp{\left(-\sum_{i=1}^{N} \ell(\boldsymbol{\theta}, \boldsymbol{x}_i) - \ell(\boldsymbol{\theta}, \boldsymbol{x}_h)\right)}}\right)} \\ 
    &= \log{\left(\frac{Z_h}{Z_l}\right)} + \left(\ell(\boldsymbol{\theta}, \boldsymbol{x}_h) - \ell(\boldsymbol{\theta}, \boldsymbol{x}_l)\right) \label{eq:diff eq2}. 
\end{align}
Introducing $\Delta(\boldsymbol{\theta}) = \ell(\boldsymbol{\theta}, \boldsymbol{x}_h) - \ell(\boldsymbol{\theta}, \boldsymbol{x}_l)$ for notational convenience. A rewrite of $Z_h$ yields, 
\begin{align}
    Z_h &= \int \exp{\left(-\sum_{i=1}^{N} \ell(\boldsymbol{\theta}, \boldsymbol{x}_i) - \ell(\boldsymbol{\theta}, \boldsymbol{x}_l)\right)}\exp{\left(\ell(\boldsymbol{\theta}, \boldsymbol{x}_l) - \ell(\boldsymbol{\theta}, \boldsymbol{x}_h)\right)}\, d\boldsymbol{\theta} \\ 
    &= \int Z_l p(\boldsymbol{\theta}\mid  \mathcal{X}_l) \exp{(-\Delta (\boldsymbol{\theta}))} \, d \boldsymbol{\theta} \\ 
    &= Z_l \mathbb{E}_{\boldsymbol{\theta} \mid  \mathcal{X}_l}[\exp{(-\Delta (\boldsymbol{\theta}))}]. 
\end{align}
Substitute back into \eqref{eq:diff eq2}: 
\begin{align}
    \log{\left(\frac{Z_h}{Z_l}\right)} + \Delta (\boldsymbol{\theta}) = \log{\left(\mathbb{E}_{\boldsymbol{\theta} \mid  \mathcal{X}_l}[\exp(-\Delta(\boldsymbol\theta))]\right)} + \Delta(\boldsymbol{\theta}),  
\end{align}
and subsequently into \eqref{eq:diff eq1}, the difference in KL divergence admits the following closed-form expression: 
\begin{align}
    \mathrm{KL}_h - \mathrm{KL}_l &= \int p(\boldsymbol{\theta} \mid \mathcal{X}) \left(\log{\left(\mathbb{E}_{\boldsymbol{\theta} \mid \mathcal{X}_l}[\exp(-\Delta(\boldsymbol\theta))]\right)} + \Delta(\boldsymbol{\theta})\right)\, d \boldsymbol{\theta}
    \\ &= \log\left(\mathbb{E}_{\boldsymbol{\theta}\mid \mathcal{X}_l}[\exp(-\Delta (\boldsymbol{\theta}))]\right) + \mathbb{E}_{\boldsymbol{\theta} \mid \mathcal{X}}[\Delta (\boldsymbol{\theta})]. 
\end{align}
The use of the cumulant generating function used in Proposition \ref{cor:var fraction} can similarly be applied here: 
\begin{align}\label{eq:diff eq3}
    \mathrm{KL}_h - \mathrm{KL}_l &\approx \frac{1}{2} \mathrm{Var}_{\boldsymbol{\theta} \mid \mathcal{X}_l}[\Delta (\boldsymbol{\theta})] - \mathbb{E}_{\boldsymbol{\theta} | \mathcal{X}_l}[\Delta (\boldsymbol{\theta})] + \mathbb{E}_{\boldsymbol{\theta} | \mathcal{X}}[\Delta (\boldsymbol{\theta})]
\end{align}
To rewrite the last term to follow the same distribution as the first two, we find the relation between the two distributions: 
\begin{align*}
    \frac{p(\boldsymbol{\theta}\mid \mathcal{X})}{p(\boldsymbol{\theta} \mid  \mathcal{X}_l)} &= \frac{Z_l}{Z} \exp{(\ell(\boldsymbol{\theta}, \boldsymbol{x}_l))} \iff \\
    p(\boldsymbol{\theta} \mid \mathcal{X}) &= \frac{Z_l}{Z} \exp{(\ell(\boldsymbol{\theta}, \boldsymbol{x}_l))} p(\boldsymbol{\theta} \mid  \mathcal{X}_l). 
\end{align*}
Integration over $\boldsymbol{\theta}$ on both sides yields $Z_l/Z=(\mathbb{E}_{\boldsymbol{\theta} \mid  \mathcal{X}_l}[\exp{(\ell (\boldsymbol{\theta}}, \boldsymbol{x}_l))])^{-1}$, such that if we multiply both sides with $\Delta(\boldsymbol{\theta})$ and integrate we get
\begin{align}
    \mathbb{E}_{\boldsymbol{\theta} \mid \mathcal{X}} [\Delta (\boldsymbol{\theta})] = \frac{\mathbb{E}_{\boldsymbol{\theta} \mid \mathcal{X}_l}[\exp{(\ell(\boldsymbol{\theta}, \boldsymbol{x}_l))}\Delta (\boldsymbol{\theta})]}{\mathbb{E}_{\boldsymbol{\theta} \mid  \mathcal{X}_l}[\exp{(\ell(\boldsymbol{\theta}, \boldsymbol{x}_l))}]}. 
\end{align}
Substitute back into \eqref{eq:diff eq3}: 
\begin{align}
    \mathrm{KL}_h - \mathrm{KL}_l &\approx  \frac{1}{2} \mathrm{Var}_{\boldsymbol{\theta} \mid \mathcal{X}_l}[\Delta (\boldsymbol{\theta})] - \mathbb{E}_{\boldsymbol{\theta} | \mathcal{X}_l}[\Delta (\boldsymbol{\theta})] + \frac{\mathbb{E}_{\boldsymbol{\theta} \mid  \mathcal{X}_l}[\exp{(\ell(\boldsymbol{\theta}, \boldsymbol{x}_l))}\Delta (\boldsymbol{\theta})]}{\mathbb{E}_{\boldsymbol{\theta} \mid  \mathcal{X}_l}[\exp{(\ell(\boldsymbol{\theta}, \boldsymbol{x}_l))}]} \\
    &= \frac{1}{2} \mathrm{Var}_{\boldsymbol{\theta} \mid \mathcal{X}_l}[\Delta (\boldsymbol{\theta})]  + \frac{\mathbb{E}_{\boldsymbol{\theta} \mid  \mathcal{X}_l}[\exp{(\ell(\boldsymbol{\theta}, \boldsymbol{x}_l))}\Delta (\boldsymbol{\theta})] - \mathbb{E}_{\boldsymbol{\theta} | \mathcal{X}_l}[\Delta (\boldsymbol{\theta})] \mathbb{E}_{\boldsymbol{\theta} \mid \mathcal{X}_l}[\exp{(\ell(\boldsymbol{\theta}, \boldsymbol{x}_l))}]}{\mathbb{E}_{\boldsymbol{\theta} \mid \mathcal{X}_l}[\exp{(\ell(\boldsymbol{\theta}, \boldsymbol{x}_l))}]}\label{eq:diff kl 41} \\
    &=  \frac{1}{2} \mathrm{Var}_{\boldsymbol{\theta} \mid \mathcal{X}_l}[\Delta (\boldsymbol{\theta})]  + \frac{\mathrm{Cov}\left[\Delta (\boldsymbol{\theta}), \exp{\left(\ell(\boldsymbol{\theta}, \boldsymbol{x}_l)\right)} \right]}{\mathbb{E}_{\boldsymbol{\theta} \mid \mathcal{X}_l}[\exp{(\ell(\boldsymbol{\theta}, \boldsymbol{x}_l))}]}\label{eq:diff to tight}
\end{align}
Using (\ref{itm:A1}) on the loss function: 
\begin{align}
    \ell(\boldsymbol{\theta}, \boldsymbol{x}_h) \approx  \ell(\boldsymbol{\theta}, \boldsymbol{x}_l) + \nabla^\top_{\boldsymbol{x}}\ell(\boldsymbol{\theta}, \boldsymbol{x})\big|_{\boldsymbol{x}=\boldsymbol{x}_l} \boldsymbol{x}_r = \ell(\boldsymbol{\theta}, \boldsymbol{x}_l) + \boldsymbol{g}^\top\boldsymbol{x}_r, 
\end{align}
introducing $\boldsymbol{g} = \nabla_{\boldsymbol{x}}\ell(\boldsymbol{\theta}, \boldsymbol{x})\big|_{\boldsymbol{x}=\boldsymbol{x}_l}$ such that $\Delta(\boldsymbol{\theta}) = \boldsymbol{g}^\top \boldsymbol{x}_r$. From this point onward in the derivation all expectations, variances and covariances are taken with respect to $P(\boldsymbol{\theta} \mid \mathcal{X}\cup \{\boldsymbol{x}_l\})$, and we will thus drop subscripts. We then obtain the following lower bound: 
\begin{align*}
    \mathrm{KL}_h - \mathrm{KL}_l
    &\approx \tfrac{1}{2} \mathrm{Var}[\boldsymbol{g}^\top \boldsymbol{x}_r] + \frac{\mathrm{Cov}[\boldsymbol{g}^\top \boldsymbol{x}_r , \exp{(\ell(\boldsymbol{\theta}, \boldsymbol{x}_l))}]}{\mathbb{E}[\exp{(\ell(\boldsymbol{\theta}, \boldsymbol{x}_l))}]} \\
    &\overset{\ref{itm:A2}}{\geq} \tfrac{1}{2} \mathrm{Var}[\boldsymbol{g}^\top \boldsymbol{x}_r] - \frac{\sqrt{\mathrm{Var}[\exp{\left(\ell(\boldsymbol{\theta}, \boldsymbol{x}_l)\right)}]}}{\mathbb{E}[\exp\left(\ell(\boldsymbol{\theta}, \boldsymbol{x}_l)\right)]} \cdot \sqrt{\mathrm{Var}[\boldsymbol{g}^\top \boldsymbol{x}_r}]  \\
    &\overset{\ref{itm:A3}+\ref{itm:A4}}{=} \tfrac{1}{2} \|\boldsymbol{x}_r\|_{\Sigma^l_{\boldsymbol{g}}}^2 - \zeta \|\boldsymbol{x}_r\|_{\Sigma^l_{\boldsymbol{g}}}, 
\end{align*}
where $\zeta=\sqrt{ \mathrm{Var}_{\boldsymbol{\theta} \mid \mathcal{X}  \cup \{\boldsymbol{x}_l\}}[\exp\left(\ell(\boldsymbol{\theta}, \boldsymbol{x}_l))\right])} /  \mathbb{E}_{\boldsymbol{\theta} \mid \mathcal{X}\cup \{\boldsymbol{x}_l\}}[\exp\left(\ell(\boldsymbol{\theta}, \boldsymbol{x}_l)\right)]$. 
The upper bound is found by upper bounding the covariance term using \eqref{itm:A2} which results in
\begin{align}
   \mathrm{KL}_h - \mathrm{KL}_l \leq  \tfrac{1}{2} \|\boldsymbol{x}_r\|_{\Sigma^l_{\boldsymbol{g}}}^2 + \zeta \|\boldsymbol{x}_r\|_{\Sigma^l_{\boldsymbol{g}}}. 
\end{align}

\section{Additional Theoretical Results}\label{app:second order}
For notational simplicity, we add some additional definitions: 
\begin{enumerate}
\item For a matrix $H\in \mathbb{R}^{s\times s}$ define $\Sigma_H:=\mathrm{Cov}[\mathrm{vec}(H)]$
    \item $Q_H(\boldsymbol{x}_r):=\left(\mathrm{vec}\left(\boldsymbol{x}_r\boldsymbol{x}_r^\top \right)^\top\Sigma_{H}\mathrm{vec}\left(\boldsymbol{x}_r\boldsymbol{x}_r^\top \right)\right)^{1/2}$
\end{enumerate}
such that 
\begin{align}\label{eq:var sigma H}
    \mathrm{Var}[\boldsymbol{x}_r^\top H \boldsymbol{x}_r] = Q_H(\boldsymbol{x}_r)^2 
\end{align}

\subsection{Second Order Approximations of Proofs}
Here we provide proofs of Proposition \ref{thm: KL ratio}-\ref{thm: Delta KL} using a second order Taylor expansion. 
\paragraph{Proposition \ref{thm: KL ratio}}
A second order Taylor expansion of the loss function yields: 
\begin{align}
    \ell(\boldsymbol{\theta}, \boldsymbol{x}_h) = \ell(\boldsymbol{\theta}, \boldsymbol{x}_l + \boldsymbol{x}_r) &\approx \ell(\boldsymbol{\theta}, \boldsymbol{x}_l) + \nabla^\top_{\boldsymbol{x}}\ell(\boldsymbol{\theta}, \boldsymbol{x})\big|_{\boldsymbol{x}=\boldsymbol{x}_l}\boldsymbol{x}_r + \tfrac{1}{2} \boldsymbol{x}_r^\top \nabla^2_{\boldsymbol{x}} \ell(\boldsymbol{\theta}, \boldsymbol{x})\big |_{\boldsymbol{x}=\boldsymbol{x}_l} \boldsymbol{x}_r \\ 
    &= \ell(\boldsymbol{\theta}, \boldsymbol{x}_l) + \boldsymbol{g}^\top \boldsymbol{x}_r + \tfrac{1}{2} \boldsymbol{x}_r^\top H \boldsymbol{x}_r, 
\end{align}
where $\boldsymbol{g} = \nabla_{\boldsymbol{x}}\ell(\boldsymbol{\theta}, \boldsymbol{x})\big|_{\boldsymbol{x}=\boldsymbol{x}_l}$ and $H := \nabla^2_{\boldsymbol{x}} \ell(\boldsymbol{\theta}, \boldsymbol{x})\big |_{\boldsymbol{x} = \boldsymbol{x}_l}$. The numerator of eq. \eqref{eq:cor3 starting point1} is thus given by, 
\begin{align*}
    \mathrm{Var}[\ell(\boldsymbol{\theta}, \boldsymbol{x}_h)] &= \mathrm{Var}[\ell(\boldsymbol{\theta}, \boldsymbol{x}_l)] + \mathrm{Var}[\boldsymbol{g}^\top\boldsymbol{x}_r] + \mathrm{Var}\left[\tfrac{1}{2}\boldsymbol{x}_r^\top H \boldsymbol{x}_r\right] \\
    &+ 2\mathrm{Cov}[\ell(\boldsymbol{\theta}, \boldsymbol{x}_l), \boldsymbol{g}^\top \boldsymbol{x}_r] + 2\mathrm{Cov}\left[\ell(\boldsymbol{\theta}, \boldsymbol{x}_l), \tfrac{1}{2}\boldsymbol{x}_r^\top H \boldsymbol{x}_r\right] + 2\mathrm{Cov}\left[\boldsymbol{g}^\top \boldsymbol{x}_r, \tfrac{1}{2} \boldsymbol{x}_r^\top H \boldsymbol{x}_r\right]. 
\end{align*}
Introducing $\sigma_l^2 := \mathrm{Var}[\ell(\boldsymbol{\theta}, \boldsymbol{x}_l)]$. Apply (\ref{itm:A3}) and using (\ref{itm:A2}) to lower bound all covariances: 
\begin{align*}
    \mathrm{Var}[\ell(\boldsymbol{\theta}, \boldsymbol{x}_h)] &\geq \sigma_l^2 + \boldsymbol{x}_r^\top \Sigma_{\boldsymbol{g}}\boldsymbol{x}_r + \tfrac{1}{4}\mathrm{vec}(\boldsymbol{x}_r\boldsymbol{x}_r^\top)^\top \Sigma_{H}\mathrm{vec}(\boldsymbol{x}_r\boldsymbol{x}_r^\top) 
    - 2 \sqrt{\sigma_l^2  \boldsymbol{x}_r^\top \Sigma_{\boldsymbol{g}}\boldsymbol{x}_r} \\&-\sqrt{\sigma_l^2 \mathrm{vec}(\boldsymbol{x}_r\boldsymbol{x}_r^\top)^\top \Sigma_{H}\mathrm{vec}(\boldsymbol{x}_r\boldsymbol{x}_r^\top)} - \sqrt{\boldsymbol{x}_r^\top \Sigma_{\boldsymbol{g}}\boldsymbol{x}_r \mathrm{vec}(\boldsymbol{x}_r\boldsymbol{x}_r^\top)^\top\Sigma_{H}\mathrm{vec}(\boldsymbol{x}_r\boldsymbol{x}_r^\top)} \\
    &\overset{(\ref{itm:A4})}{=} \sigma_l^2 + \|\boldsymbol{x}_r\|^2_{\Sigma_{\boldsymbol{g}}} + \tfrac{1}{4}Q_H(\boldsymbol{x}_r)^2
    \\&- 2\sigma_l \|\boldsymbol{x}_r\|_{\Sigma_{\boldsymbol{g}}} - \sigma_l Q_H(\boldsymbol{x}_r) - \|\boldsymbol{x}_r\|_{\Sigma_{\boldsymbol{g}}} Q_H(\boldsymbol{x}_r). 
\end{align*}
Rearranging gives us the final lower bound: 
\begin{align}
    \frac{\mathrm{KL}_h}{\mathrm{KL}_l} \geq \frac{\left(\sigma_l - \|\boldsymbol{x}_r\|_{\Sigma_{\boldsymbol{g}}}\right)^2 + \tfrac{1}{4}Q_H(\boldsymbol{x}_r)^2 - \left(\sigma_l Q_H(\boldsymbol{x}_r) + \|\boldsymbol{x}_r\|_{\Sigma_{\boldsymbol{g}}} Q_H(\boldsymbol{x}_r)\right)}{\sigma_l^2}. 
\end{align}
Using (\ref{itm:A2}) to upper bound all covariances results in: 
\begin{align}
    \frac{\mathrm{KL}_h}{\mathrm{KL}_l} \leq \frac{\left(\sigma_l + \|\boldsymbol{x}_r\|_{\Sigma_{\boldsymbol{g}}}\right)^2 + \tfrac{1}{4}Q_H(\boldsymbol{x}_r)^2 + \left(\sigma_l Q_H(\boldsymbol{x}_r) + \|\boldsymbol{x}_r\|_{\Sigma_{\boldsymbol{g}}} Q_H(\boldsymbol{x}_r)\right)}{\sigma_l^2}. 
\end{align}

\paragraph{Proposition \ref{thm: Delta KL}}
A second order Taylor expansion of the loss function yields: 
\begin{align}
    \ell(\boldsymbol{\theta}, \boldsymbol{x}_h) = \ell(\boldsymbol{\theta}, \boldsymbol{x}_l + \boldsymbol{x}_r) &\approx \ell(\boldsymbol{\theta}, \boldsymbol{x}_l) + \nabla^\top_{\boldsymbol{x}}\ell(\boldsymbol{\theta}, \boldsymbol{x})\big|_{\boldsymbol{x}=\boldsymbol{x}_l} \boldsymbol{x}_r + \tfrac{1}{2} \boldsymbol{x}_r^\top \nabla^2_{\boldsymbol{x}} \ell(\boldsymbol{\theta}, \boldsymbol{x})\big |_{\boldsymbol{x}=\boldsymbol{x}_l} \boldsymbol{x}_r \\ 
    &= \ell(\boldsymbol{\theta}, \boldsymbol{x}_l) + \boldsymbol{g}^\top \boldsymbol{x}_r + \tfrac{1}{2} \boldsymbol{x}_r^\top H \boldsymbol{x}_r, 
\end{align}
where $\boldsymbol{g} = \nabla_{\boldsymbol{x}}\ell(\boldsymbol{\theta}, \boldsymbol{x})\big|_{\boldsymbol{x}=\boldsymbol{x}_l}$ and $H := \nabla^2_{\boldsymbol{x}} \ell(\boldsymbol{\theta}, \boldsymbol{x})\big |_{\boldsymbol{x}=\boldsymbol{x}_l}$. Dropping subscripts we can rewrite \eqref{eq:diff kl 41}
\begin{align*}
    \mathrm{KL}_h - \mathrm{KL}_l &\approx \tfrac{1}{2} \mathrm{Var}[\Delta(\boldsymbol{\theta})] - \mathbb{E}[\Delta (\boldsymbol{\theta})] + \frac{\mathbb{E}[\exp{(\ell(\boldsymbol{\theta}, \boldsymbol{x}_l)) \Delta(\boldsymbol{\theta})}]}{\mathbb{E}[\exp{(\ell(\boldsymbol{\theta}, \boldsymbol{x}_l))}]} \\
    &= \tfrac{1}{2} \mathrm{Var}\left[\boldsymbol{g}^\top \boldsymbol{x}_r + \tfrac{1}{2} \boldsymbol{x}_r^\top H \boldsymbol{x}_r\right] -(\mathbb{E}\left[\boldsymbol{g}^\top \boldsymbol{x}_r + \tfrac{1}{2} \boldsymbol{x}_r^\top H\boldsymbol{x}_r\right] )
    + \frac{\mathbb{E}[\exp{(\ell(\boldsymbol{\theta}, \boldsymbol{x}_l))}(\boldsymbol{g}^\top \boldsymbol{x}_r + \tfrac{1}{2} \boldsymbol{x}_r ^\top H \boldsymbol{x}_r)]}{\mathbb{E}[\exp{(\ell(\boldsymbol{\theta}, \boldsymbol{x}_l))}]}. 
\end{align*}
Applying (\ref{itm:A3}): 
\begin{align*}
     \mathrm{KL}_h - \mathrm{KL}_l &\approx \tfrac{1}{2}\|\boldsymbol{x}_r\|_{\Sigma_{\boldsymbol{g}}}^2 + \tfrac{1}{8} Q_H(\boldsymbol{x}_r)^2 + \tfrac{1}{2} \mathrm{Cov}[\boldsymbol{g}^\top \boldsymbol{x}_r, \boldsymbol{x}_r^\top H \boldsymbol{x}_r] - \mathbb{E}[\boldsymbol{g}^\top \boldsymbol{x}_r]  \\
     &- \tfrac{1}{2} \mathbb{E}[\boldsymbol{x}_r^\top H \boldsymbol{x}_r] + \frac{\mathbb{E}[\exp{(\ell(\boldsymbol{\theta}, \boldsymbol{x}_l))\boldsymbol{g}^\top \boldsymbol{x}_r}] - \frac{1}{2}\mathbb{E}[\exp{(\ell(\boldsymbol{\theta}, \boldsymbol{x}_l)) \boldsymbol{x}_r^\top H \boldsymbol{x}_r}]}{\mathbb{E}[\exp{(\ell(\boldsymbol{\theta}, \boldsymbol{x}_l))}]}. 
\end{align*}
By lower bounding the first covariance term using (\ref{itm:A2}), we can rewrite the first three terms: 
\begin{align*}
     \mathrm{KL}_h - \mathrm{KL}_l &\geq \tfrac{1}{2}\left(\|\boldsymbol{x}_r\|_{\Sigma_{\boldsymbol{g}}} - \tfrac{1}{2}Q_H(\boldsymbol{x}_r)\right)^2 +\frac{\mathbb{E}[\exp{(\ell(\boldsymbol{\theta}, \boldsymbol{x}_l))}\boldsymbol{g}^\top \boldsymbol{x}_r] - \mathbb{E}[\exp{(\ell(\boldsymbol{\theta}, \boldsymbol{x}_l))}] \mathbb{E}[\boldsymbol{g}^\top \boldsymbol{x}_r]}{\mathbb{E}[\exp{(\ell(\boldsymbol{\theta}, \boldsymbol{x}_l))}]} \\
     &+ \frac{\frac{1}{2} \mathbb{E}[\exp{(\ell(\boldsymbol{\theta}, \boldsymbol{x}_l)) \boldsymbol{x}_r^\top H \boldsymbol{x}_r}] - \frac{1}{2}\mathbb{E}[\boldsymbol{x}_r^\top H \boldsymbol{x}_r] \mathbb{E}[\exp{(\ell(\boldsymbol{\theta}, \boldsymbol{x}_l))}] }{\mathbb{E}[\exp{(\ell(\boldsymbol{\theta}, \boldsymbol{x}_l))}]} \\
     &= \tfrac{1}{2} \left(\|\boldsymbol{x}_r\|_{\Sigma_{\boldsymbol{g}}} - \tfrac{1}{2} Q_H(\boldsymbol{x}_r)\right)^2 + \frac{\mathrm{Cov}\left[\boldsymbol{g}^\top \boldsymbol{x}_r, \exp{(\ell(\boldsymbol{\theta}, \boldsymbol{x_l}))}\right]}{\mathbb{E}[\exp{(\ell(\boldsymbol{\theta}, \boldsymbol{x}_l))}]} \\
     &+ \frac{\tfrac{1}{2}\mathrm{Cov}\left[\boldsymbol{x}_r^\top H \boldsymbol{x}_r, \exp{(\ell(\boldsymbol{\theta}, \boldsymbol{x}_l))}\right]}{\mathbb{E}[\exp{(\ell(\boldsymbol{\theta}, \boldsymbol{x}_l))}]}. 
\end{align*}
Applying (\ref{itm:A2}) to lower bound the other covariances and (\ref{itm:A4}) to rewrite the norms: 
\begin{align*}
    \mathrm{KL}_h - \mathrm{KL}_l &\geq \tfrac{1}{2} \left(\|\boldsymbol{x}_r\|_{\Sigma_{\boldsymbol{g}}} - \tfrac{1}{2}Q_H(\boldsymbol{x}_r)\right)^2\\
    &- \frac{\|\boldsymbol{x}_r\|_{\Sigma_{\boldsymbol{g}}} \sqrt{\mathrm{Var}[\exp{(\ell(\boldsymbol{\theta}, \boldsymbol{x}_l))}]}}{\mathbb{E}[\exp{(\ell(\boldsymbol{\theta}, \boldsymbol{x}_l))}]} - \frac{\tfrac{1}{2} Q_H(\boldsymbol{x}_r) \sqrt{\mathrm{Var}[\exp{(\ell(\boldsymbol{\theta}, \boldsymbol{x}_l))}]}}{\mathbb{E}[\exp{(\ell(\boldsymbol{\theta}, \boldsymbol{x}_l))}]}, 
\end{align*}
which gives us the following lower bound: 
\begin{align}
    \mathrm{KL}_h - \mathrm{KL}_l &\geq\tfrac{1}{2} \left(\|\boldsymbol{x}_r\|_{\Sigma_{\boldsymbol{g}}} - \tfrac{1}{2}Q_H(\boldsymbol{x}_r)\right)^2 - c\left(\|\boldsymbol{x}_r\|_{\Sigma_{\boldsymbol{g}}} + \tfrac{1}{2} Q_H(\boldsymbol{x}_r)\right), 
\end{align}
where $c=\sqrt{\mathrm{Var}_{\boldsymbol{\theta} \mid \mathcal{X} \cup\{\boldsymbol{x}_l\}}[\exp{(\ell(\boldsymbol{\theta}, \boldsymbol{x}_l))}]} / \mathbb{E}[\exp{(\ell(\boldsymbol{\theta}, \boldsymbol{x}_l))}]$. 
Had (\ref{itm:A2}) been used to upper bound all covariances, we end up with the following upper bound: 
\begin{align}
    \mathrm{KL}_h - \mathrm{KL}_l &\leq\tfrac{1}{2} \left(\|\boldsymbol{x}_r\|_{\Sigma_{\boldsymbol{g}}} + \tfrac{1}{2}Q_H(\boldsymbol{x}_r)\right)^2 + c\left(\|\boldsymbol{x}_r\|_{\Sigma_{\boldsymbol{g}}} + \tfrac{1}{2} Q_H(\boldsymbol{x}_r)\right), 
\end{align}

\begin{figure}
    \centering
    \begin{subfigure}[b]{0.49\textwidth}
        \includegraphics[trim=0 0.2cm 0 0.2cm, clip=true]{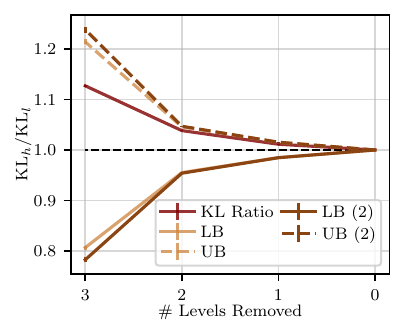}
    \end{subfigure}
    \begin{subfigure}[b]{0.49\textwidth}
        \includegraphics[trim=0 0.2cm 0 0.2cm, clip=true]{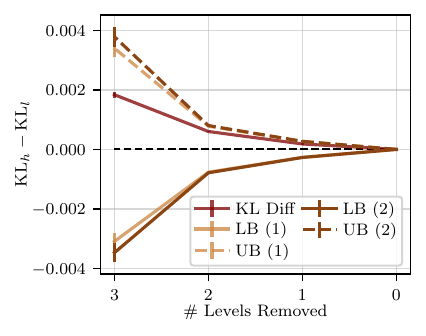}
        
    \end{subfigure}
    \caption{Simulation of bounds as a function of progressively removed high-frequency content. Error bars again denote five standard deviations. LB = Lower Bound, UB = Upper Bound with the order of approximation in parenthesis. There is no added benefit of going from a first to a second order Taylor expansion.}
    \label{fig:simulation appendix}
\end{figure}

Figure \ref{fig:simulation appendix} augments Figure \ref{fig:simulation} with the proofs using the second order approximations. There is not an added benefit of using a second order Taylor expansion for either bound. 

\subsection{Proof of Corollary \ref{cor:tight bounds}}\label{sec:tigher lower bounds}
The tighter lower bound for Proposition \ref{thm: KL ratio} follows directly from \eqref{eq:p5 low bound} lower-bounding the covariance with zero. 

Likewise the tighter lower bound for Proposition \ref{thm: Delta KL} follows from lower bounding the covariance in \eqref{eq:diff to tight} with zero. 

\subsection{Proof of Proposition \ref{thm: lin reg}}\label{sec:lin reg proof}
The loss function used in linear regression is the Mean Squared Error (MSE) which is defined as
\begin{align}
    MSE(\boldsymbol{\theta}; \boldsymbol{x}, y):=(y - \boldsymbol{x}^\top \boldsymbol{\theta})^2. 
\end{align}
Inserting this into Gibbs posterior from Assumption \ref{ass:gibbs}: 
\begin{align}
    p(\boldsymbol{\theta} \mid \mathcal{X}) = \frac{1}{Z(\mathcal{X})} \exp{\left(-\sum_{i=1}^N (y_i - \boldsymbol{x}_i^\top \boldsymbol{\theta})^2\right)} = \mathcal{N}(\boldsymbol{\theta} \mid \boldsymbol{\mu}, \Sigma), 
\end{align}
thus $\boldsymbol{\theta} \sim \mathcal{N}(\boldsymbol{\mu}, \boldsymbol{\Sigma})$. Introducing $z_i:=\boldsymbol{x}_i^\top \boldsymbol{\theta} \sim \mathcal{N}(m, s^2)$ where \citep{murphy2022probabilistic}
\begin{align}
    m = \boldsymbol{x}_i^\top \boldsymbol{\mu}, \qquad s^2 = \boldsymbol{x}_i^\top \boldsymbol{\Sigma} \boldsymbol{x}_i. 
\end{align}
Insert the definition of $z_i$ in the exponential term from equation  \ref{lem:exact KL}:  
\begin{align*}
    \mathbb{E}_{\boldsymbol{\theta} \mid \mathcal{X}}[\exp{\left(-(y_i - \boldsymbol{x}_i^\top \boldsymbol{\theta})^2\right)}]  &= \mathbb{E}_{z_i \sim \mathcal{N}(m, s^2)}[\exp{(-(y_i - z_i)^2)}] \\
    &= \int \exp{\left(-(y_i - z_i)^2\right)} \frac{1}{\sqrt{2\pi s^2}} \exp{\left(-\frac{(z_i-m)^2}{2s^2}\right)}\, d z_i \\
    &= \frac{1}{\sqrt{2\pi s^2}} \int \exp{\left(-(y_i-z_i)^2 - \frac{(z_i - m)^2}{2s^2}\right)}\, dz_i \label{eq:mse exponent}. 
\end{align*}
Rewriting the exponent: 
\begin{align*}
    (y-z_i)^2 + \frac{(z_i-m)^2}{2s^2} &= y^2 + z_i^2 - 2yz_i +\frac{z_i^2 + m^2 - 2z_im}{2s^2} \\
    &= az_i^2 + bz_i + c, 
\end{align*}
where $a=1 + 1/(2s^2)$, $b=-2y - m/s^2$ and $c = y^2 + m^2/(2s^2)$. The exponent can thus be rewritten to
\begin{align*}
    -(az_i^2 + bz_i + c) &= -a \left(z_i + \frac{b}{2a}\right)^2 + \frac{b^2}{4a} - c \\
    &= -a'(z_i - b')^2 + c', 
\end{align*}
where $a' = a$, $b' = -b/(2a)$ and $c'=b^2/(4a) - c$. Substitute back into the exponential term: 
\begin{align*}
     \mathbb{E}_{\boldsymbol{\theta} \mid \mathcal{X}}[\exp{\left(-(y_i - \boldsymbol{x}_i^\top \boldsymbol{\theta})^2\right)}] &= \frac{1}{\sqrt{2\pi s^2}} \int \exp{\left(-a' (z_i - b')^2+c'\right) }\, d z_i \\
    &= \frac{1}{\sqrt{2\pi s^2}} \exp(c')\int \exp{\left(-a' (z_i - b')^2\right) }\, d z_i \\
    &= \frac{1}{\sqrt{2\pi s^2}} \exp{(c')} \sqrt{\frac{\pi}{a'}}
\end{align*}
Inserting the definitions of $c'$ and $a'$: 
\begin{align*}
     \mathbb{E}_{\boldsymbol{\theta} \mid \mathcal{X}}[\exp{\left(-(y_i - \boldsymbol{x}_i^\top \boldsymbol{\theta})^2\right)}] &= \frac{\sqrt{\pi}}{\sqrt{2\pi s^2 \left(1+\frac{1}{2s^2}\right)}} \exp{\left(\frac{b^2}{4a} - c\right)} \\
     &= \frac{1}{\sqrt{2s^2 + 1}} \exp{\left(\frac{(-2y_i-\frac{m}{s^2})^2}{4a} - (y_i^2+\frac{m^2}{2s^2})\right)} \\
     &= \frac{1}{\sqrt{2s^2+1}} \exp{\left(\frac{4y^2+\frac{m^2}{s^4} + \frac{4y_im}{s^2}}{4a} - (y_i^2 + \frac{m^2}{2s^2})\right)} \\
     &= \frac{1}{\sqrt{2s^2+1}}  \exp{\left(\frac{1}{a}\left(y_i^2 + \frac{m^2}{4s^4} + \frac{y_im}{s^2}\right) - (y_i^2 + \frac{m^2}{2s^2})\right)}. 
\end{align*}
Since $1/a = 2s^2/(2s^2+1)$ then the exponent, $c'$ can be rewritten as such: 
\begin{align*}
    c' &= \frac{2s^2}{2s^2+1} \left(y_i^2 + \frac{m^2}{4s^4} + \frac{y_im}{s^2}\right) - \left(y_i^2 + \frac{m^2}{2s^2}\right) \\
    &= \frac{2s^2y_i^2}{2s^2+1} + \frac{m^2}{(2s^2+1)2s^2} + \frac{2my_i}{1+2s^2} - y_i^2 - \frac{m^2}{2s^2} \\
    &= y_i^2 \left(\frac{2s^2}{2s^2+1} - 1\right) + \frac{2my}{2s^2+1} + \frac{m^2}{2s^2} \left(\frac{1}{2s^2+1} - 1\right) \\
    &= -\frac{y_i^2}{2s^2+1} + \frac{2y_im}{2s^2+1}  - \frac{m^2}{2s^2+1} \\
    &= \frac{1}{2s^2+1} \left(-y_i^2-m^2 + 2y_im\right) \\
    &=- \frac{\left(y_i-m\right)^2}{2s^2+1}, 
\end{align*}
and thus 
\begin{align*}
     \mathbb{E}_{\boldsymbol{\theta} \mid \mathcal{X}}[\exp{\left(-(y_i - \boldsymbol{x}_i^\top \boldsymbol{\theta})^2\right)}] &= \frac{1}{\sqrt{1+2s^2}} \exp{\left(-\frac{(y_i - m)^2}{2s^2+1}\right)} \Rightarrow \\ 
     \log   \mathbb{E}_{\boldsymbol{\theta} \mid \mathcal{X}}[\exp{\left(-(y_i - \boldsymbol{x}_i^\top \boldsymbol{\theta})^2\right)}] &=-\frac{1}{2}\log(1+2s^2) - \frac{(y_i - m)^2}{2s^2+1}. 
\end{align*}
Computing the other term from \eqref{lem:exact KL}: 
\begin{align*}
    \mathbb{E}_{\boldsymbol{\theta} \mid \mathcal{X}}[\ell(\boldsymbol{\theta}, \boldsymbol{x}_i)] &= \mathbb{E}_{\boldsymbol{\theta} \mid \mathcal{X}}[(y_i - \boldsymbol{x}_i^\top \boldsymbol{\theta})^2] \\
    &= \mathbb{E}_{z_i \sim \mathcal{N}(m, s^2)}[(y_i - z_i)^2] \\
    &= y_i^2 + \mathbb{E}_{z_i \sim \mathcal{N}(m, s^2)}[z_i^2] - 2 y_i \mathbb{E}_{z_i \sim \mathcal{N}(m, s^2)}[z_i] \\
    &= y_i^2 + m^2 + s^2 -2y_i m \\
    &= (y_i-m)^2 + s^2. 
\end{align*}
$\mathrm{KL}_i$ can thus be written as 
\begin{align}
    \mathrm{KL}_i =
    & -\frac{1}{2}\log(1+2s^2) - \frac{(y_i - m)^2}{2s^2+1} + (y_i-m)^2 + s^2
\end{align}
We have for different limit values of $s^2$: 
\begin{align}
    \mathrm{KL}_i \to \begin{cases}
        0 \ &\text{for } s^2 \to 0 \\
        \infty \ &\text{for } s^2 \to \infty 
    \end{cases},
\end{align}
which means the influence of a datapoint $\boldsymbol{x}_i$ can be expressed as such: 
\begin{align}
    \mathrm{KL}_i = \mathcal{O}(s^2), 
\end{align}
for the case of linear regression. This aligns well with Proposition \ref{cor:var fraction}.

\subsection{Analytical Covariance Assumption}\label{app:lin reg cov ass}
The covariance assumption for the fraction of KL-divergences looks the following:  
\begin{align}
    \mathrm{Cov}_{\boldsymbol{\theta} \mid \mathcal{X}}[\boldsymbol{g}^\top \boldsymbol{x}_r, \ell(\boldsymbol{\theta}, \boldsymbol{x}_l)] \geq 0.  
\end{align}
First we derive $\boldsymbol{g}$: 
\begin{align}
    \boldsymbol{g} &:= \nabla_{\boldsymbol{x}} \ell(\boldsymbol{\theta}, \boldsymbol{x}) \big |_{\boldsymbol{x} = \boldsymbol{x}_l} = \nabla_{\boldsymbol{x}} (y - \boldsymbol{x}^\top \boldsymbol{\theta})^2 \big |_{\boldsymbol{x}=\boldsymbol{x}_l} \\
    &= -2(y - \boldsymbol{x}_l^\top \boldsymbol{\theta})\boldsymbol{\theta}. 
\end{align}
Inserting in the covariance, and introducing $z_i:=\boldsymbol{x}_i^\top \boldsymbol{\theta}$
\begin{align*}
     \mathrm{Cov}_{\boldsymbol{\theta} \mid \mathcal{X}}[\boldsymbol{g}^\top \boldsymbol{x}_r, \ell(\boldsymbol{\theta}, \boldsymbol{x}_l)] &= \mathrm{Cov}_{\boldsymbol{\theta} \mid \mathcal{X}}[-(2(y - \boldsymbol{x}_l^\top \boldsymbol{\theta})\boldsymbol{\theta})^\top \boldsymbol{x}_r, (y - \boldsymbol{x}_l^\top \boldsymbol{\theta})^2] \\
     &=\mathrm{Cov}_{z_l \sim \mathcal{N}(m, s^2)}[-2(y - z_l)\boldsymbol{\theta}^\top \boldsymbol{x}_r, (y - z_l)^2] \\
     &= \mathrm{Cov}_{(z_l,z_r)\sim \mathcal{N}(\boldsymbol{m},\Sigma_{lr})}[-2(y - z_l)z_r, (y - z_l)^2]. 
\end{align*}
Introduce the residual $r_i:= y - z_i$: 
\begin{align*}
     \mathrm{Cov}_{\boldsymbol{\theta} \mid \mathcal{X}}[\boldsymbol{g}^\top \boldsymbol{x}_r, \ell(\boldsymbol{\theta}, \boldsymbol{x}_l)] &= -2 \mathrm{Cov}_{(z_l,z_r)\sim \mathcal{N}(\boldsymbol{m},\Sigma_{lr})}[r_lz_r, r_l^2] \\
     &= -2 \left(\mathbb{E}_{(z_l, z_r) \sim \mathcal{N}(\boldsymbol{m}, \Sigma_{lr})}[r_l^3 z_r] - \mathbb{E}_{(z_l, z_r)\sim \mathcal{N}(\boldsymbol{m}, \Sigma_{lr})}[r_lz_r]\mathbb{E}_{z_l\sim \mathcal{N}(m, s^2)}[r_l^2]\right). 
\end{align*}
Now we have these three terms that we will look at one-by-one: 
\begin{enumerate}
    \item $\mathbb{E}_{z_l \sim \mathcal{N}(m, s^2)}[r_l^2]$ has been derived above, as this is just the expected loss: 
    \begin{align*}
        \mathbb{E}_{z_l \sim \mathcal{N}(m, s^2)}[r_l^2] = (y - m_l)^2 + s_l^2 
    \end{align*}
    \item $\mathbb{E}_{(z_l, z_r) \sim \mathcal{N}(m, \Sigma_{lr})}[r_lz_r]$ use the rule about the product about distribution of two normally distributed variables 
    \begin{align*}
        \mathbb{E}_{(z_l, z_r) \sim \mathcal{N}(\boldsymbol{m}, \Sigma_{lr})}[r_lz_r] &= \mathbb{E}[y z_r] - \mathbb{E}[z_lz_r] \\
        &= y m_r - (m_l m_r+ \boldsymbol{x}_l^\top \Sigma_{\boldsymbol{\theta}} \boldsymbol{x}_r) \\ 
        &= (y - m_l) m_r - \boldsymbol{x}_l^\top\Sigma_{\boldsymbol{\theta}} \boldsymbol{x}_r 
    \end{align*}
    \item $\mathbb{E}_{(z_l, z_r) \sim \mathcal{N}(\boldsymbol{m}, \Sigma_{\boldsymbol{\theta}})}[r_l^3z_r]$. Introduce $u=z_l - m_l$, $v=z_r - m_r$ and $a = y - m_l$. Then
    \begin{align*}
        \mathbb{E}_{(z_l, z_r) \sim \mathcal{N}(\boldsymbol{m}, \Sigma_{lr})}[r_l^3z_r] &= \mathbb{E}[(y - (m_l+u))^3 (m_r + v)] \\
        &= \mathbb{E}[(a - u)^3(m_r+v)] \\
        &= \mathbb{E}[(a^3 - 3a^2u+3au^2 - u^3)(m_r+v)] \\
        &= \mathbb{E}[a^3m_r -3a^2um_r + 3au^2m_r -u^3 mr \\
        & + a^3 v - 3a^2uv + 3au^2v - u^3v] \\
        &= a^3m_r + 3as_l^2 m_r - 3a^2 \mathbb{E}[uv] - \mathbb{E}[u^3v] \\
        &= a^3m_r + 3as_l^2 m_r - 3a^2 \boldsymbol{x}_l^\top \Sigma_{\boldsymbol{\theta}} \boldsymbol{x}_r - 3s_l^2 \boldsymbol{x}_l^\top \Sigma_{\boldsymbol{\theta}} \boldsymbol{x}_r \\
        &= m_r((y - m_l)^3 + 3(y - m_l)s_l^2) - 3\boldsymbol{x}_l^\top \Sigma_{\boldsymbol{\theta}}\boldsymbol{x}_r((y-m_l)^2 + s_l^2)
    \end{align*}
\end{enumerate}
Joining the three terms: 
\begin{align*}
    \mathrm{Cov}_{\boldsymbol{\theta} \mid \mathcal{X}} [\boldsymbol{g}^\top \boldsymbol{x}_r, \ell(\boldsymbol{\theta}, \boldsymbol{x}_l)] &= -2 \left(\mathbb{E}_{(z_l, z_r) \sim \mathcal{N}(\boldsymbol{m}, \Sigma_{lr})}[r_l^3 z_r] - \mathbb{E}_{(z_l, z_r)\sim \mathcal{N}(\boldsymbol{m}, \Sigma_{lr})}[r_lz_r]\mathbb{E}_{z_l\sim \mathcal{N}(m, s^2)}[r_l^2]\right) \\
    &= -2\big(m_r((y - m_l)^3 + 3(y - m_l)s_l^2) -3 \boldsymbol{x}_l^\top\Sigma_{\boldsymbol{\theta}} \boldsymbol{x}_r ((y - m_l)^2 + s_l^2) \\
    &- ((y - m_l)^2 + s_l^2) \cdot((y - m_l)m_r - \boldsymbol{x}_l^\top \Sigma_{\boldsymbol{\theta}} \boldsymbol{x}_r))\big) \\
    &= -2\big( m_r(y - m_l)^3 + 3m_r (y - m_l)s_l^2 - 3 \boldsymbol{x}_l^\top \Sigma_{\boldsymbol{\theta}} \boldsymbol{x}_r (y - m_l)^2 - 3\boldsymbol{x}_l^\top \Sigma_{\boldsymbol{\theta}} \boldsymbol{x}_r  \\
    &- (y - m_l)^3 m_r + \boldsymbol{x}_l^\top \Sigma_{\boldsymbol{\theta}}\boldsymbol{x}_r (y - m_l)^2 - s_l^2 (y - m_l)m_r + s_l^2 \boldsymbol{x}_l^\top \Sigma_{\boldsymbol{\theta}} \boldsymbol{x}_r \big). 
\end{align*}
Group terms by $m_r$ and $\Sigma_{\boldsymbol{\theta}}$: 
\begin{align*}
     \mathrm{Cov}_{\boldsymbol{\theta} \mid \mathcal{X}} [\boldsymbol{g}^\top \boldsymbol{x}_r, \ell(\boldsymbol{\theta}, \boldsymbol{x}_l)] &= -2 \bigg(m_r \big((y - m_l)^3 + 3(y - m_l)s_l^2 - (y - m_l)^3 - s_l^2(y - m_l)\big) \\
     &+ \boldsymbol{x}_l^\top\Sigma_{\boldsymbol{\theta}} \boldsymbol{x}_r \big( -3 (y - m_l)^2 - 3s_l^2 + (y - m_l)^2 + s_l^2\big) \bigg)  \\
     &= -2\big(2m_r (y - m_l)s_l^2 - \boldsymbol{x}_l^\top \Sigma_{\boldsymbol{\theta}} \boldsymbol{x}_r (2(y - m_l)^2+ 2s_l^2)\big), 
\end{align*}
hence the covariance is given by 
\begin{align}\label{eq:linreg cov ass}
    \mathrm{Cov}_{\boldsymbol{\theta} \mid \mathcal{X}}[\boldsymbol{g}^\top \boldsymbol{x}_r, \ell(\boldsymbol{\theta}, \boldsymbol{x}_l)] = -4 m_r (y - m_l)s_l^2 + 4 \boldsymbol{x}_l^\top \Sigma_{\boldsymbol{\theta}} \boldsymbol{x}_r((y - m_l)^2 + s_l^2). 
\end{align}
That means in order for the assumption to hold, the following must be true: 
\begin{align}
    \boldsymbol{x}_l^\top \Sigma_{\boldsymbol{\theta}} \boldsymbol{x}_r((y - m_l)^2 + s_l^2) \geq  m_r (y - m_l)s_l^2. 
\end{align}

In order to understand this covariance assumption better, we conduct some simulations. That requires we generate some data that can be considered of high- and low-resolution in the linear regression case. 
\subsubsection{Generate Data for Linear Regression}
To study when the covariance condition holds, we require high- and low-resolution versions of the same datapoint in a linear regression setting. For each experiment, we generate 200 signals of length 20 and fit a linear regession model with a bias term and one weight per coefficient. Signals are generated in the frequency domain by sampling coefficients from a normal distribution and scaling them by 
\begin{align}
    h(f)=\frac{1}{(1+f)^{\alpha}}, \quad \alpha=2, 
\end{align}
where $f$ is frequency index, before being transformed back to the signal domain. Low-resolution samples are obtained by low-pass filtering with a Kaiser window. The scaling of the high-frequencies ensures that a moderate amount of energy is removed by the low-pass filter and that label information is not too concentrated in high-frequency components. Targets are generated as $\boldsymbol{y} = \boldsymbol{X\theta}^* + \boldsymbol{\eta}$, where $\boldsymbol{\theta}^*$ is sampled from a normal distribution and $\eta \sim \mathcal{N}(0, 0.5^2)$. From the resulting Gaussian posterior presented in Proposition \ref{thm: lin reg}, we compute $m_l$, $m_r$, $s_l^2$ and $\boldsymbol{x}_l^\top \Sigma_{\boldsymbol{\theta}} \boldsymbol{x}_r$ for each datapoint. The procedure is repeated 2000 times.

\subsubsection{Plotting the Covariance Simulations}
We begin with rewriting \eqref{eq:linreg cov ass} to make it dependent on fewer parameters: 
\begin{align}
     \mathrm{Cov}_{\boldsymbol{\theta} \mid \mathcal{X}}[\boldsymbol{g}^\top \boldsymbol{x}_r, \ell(\boldsymbol{\theta}, \boldsymbol{x}_l)] &= -4 m_r (y - m_l)s_l^2 + 4 \boldsymbol{x}_l^\top \Sigma_{\boldsymbol{\theta}} \boldsymbol{x}_r((y - m_l)^2 + s_l^2) \iff \\
     \frac{\mathrm{Cov}[\boldsymbol{g}^\top \boldsymbol{x}_r, \ell(\boldsymbol{\theta}, \boldsymbol{x}_l)]}{4s_l^4} &= \rho(u^2 +1) - \nu u\label{eq:app cov rewrite}, 
\end{align}
where 
\begin{align}
    \rho = \frac{\boldsymbol{x}_l^\top \Sigma_{\boldsymbol{\theta}}\boldsymbol{x}_r}{s_l^2}, \qquad \nu = \frac{m_r}{s_l}, \qquad u = \frac{(y - m_l)}{s_l}.
\end{align}
Further introducing 
\begin{align}
    r = \frac{\nu}{\rho} = \frac{m_rs_l}{\boldsymbol{x}_l^\top \Sigma_{\boldsymbol{\theta}}\boldsymbol{x}_r}, 
\end{align}
such that 
\begin{align}
    \frac{\mathrm{Cov}[\boldsymbol{g}^\top \boldsymbol{x}_r, \ell(\boldsymbol{\theta}, \boldsymbol{x}_l)]}{4s_l^4} &= \rho (u^2 + 1) - \rho r u \\
    &= \rho(u^2+1-ru). 
\end{align}
The sign of the covariance is therefore given by
\begin{align}
    \mathrm{sign}(\mathrm{Cov}[\boldsymbol{g}^\top \boldsymbol{x}_r, \ell(\boldsymbol{\theta}, \boldsymbol{x}_l)]) = \mathrm{sign}(\rho) \mathrm{sign}(u^2 -ru+1). 
\end{align}
The sign changes where $u^2-ru+1=0$ i.e. where $r=u+1/u$. Since $|u+1/u| \geq2$, then the polynomial is strictly positive when $|r| < 2$. In this region, the sign of the covariance, is determined by $\rho$ alone, and the condition holds if and only if $\boldsymbol{x}_l^\top \Sigma_{\boldsymbol{\theta}}\boldsymbol{x}_r \geq 0$. When $|r| > 2$, the polynomial is negative for $u$ between the roots $(r \pm \sqrt{r^2 - 4}) / 2$, so the sign flips within this band. 

The two cases are most easily interpreted directly from \eqref{eq:app cov rewrite}, that shows how the assumption holds if and only if $\rho(u^2+1) \geq \nu u$. The two terms reflect two ways in which the residual component can correct the low-resolution prediction. The term $\nu u$ is positive when the mean residual prediction $m_r$ has the same sign as the low-resolution error, $y - m_l$, i.e. when including $\boldsymbol{x}_r$ moves the mean prediction towards the target. The term $\rho$ is negative when the predictions of the low-resolution and the residual components are negatively correlated. The assumption is therefore violated when the residual component corrects the low-resolution mean prediction or its uncertainty (given by $\rho$). In these cases the high-resolution datapoint can have a lower loss variance than the low-resolution counterpart. 

\begin{figure}
    \centering
    \includegraphics{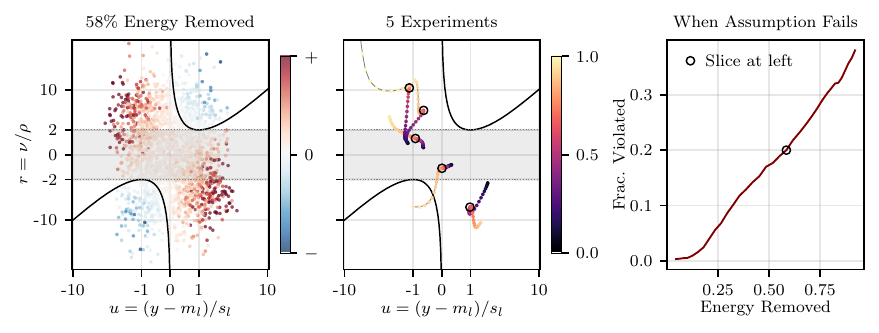}
    \caption{The covariance condition of Corollary \ref{cor:tight bounds} for linear regression. \textbf{Left}: datapoints at $58\%$ energy removed, plotted by normalised residual $u = (y - m_l)/s_l$ and $r = \nu/\rho$, coloured by $\operatorname{Cov}/(4s_l^4)$.  The black curve $r = u + 1 / u$ marks where the sign flips. \textbf{Middle}: trajectories of five datapoints as energy is removed with black corresponding to not energy removal. The black circle marks the slice seen on the left. \textbf{Right}: fraction of datapoints violating the condition as a function of energy removed; the circle marks the slice shown on the left.}
    \label{fig:covariance assumotion}
\end{figure}

Figure \ref{fig:covariance assumotion} visualises these regimes for the synthetic data described in the previous section. The left panel shows $r$ against $u$ at $58\%$ energy removed, together with the curve $r=u+1/u$, with points coloured by the normalised covariance. The middle panel follows five datapoints as energy is progressively removed, and the right panel shows the fraction of datapoints violating the assumption as a function of removed energy. Violations become more frequent as more energy is discarded, remaining below $20\%$ up to roughly $58\%$ energy removed and reaching about $35\%$ at the highest levels. When the assumption is violated, only the tighter bounds of Corollary \ref{cor:tight bounds} cease to apply, but the bounds from Proposition \ref{thm: KL ratio} still hold up. 

\section{Simulation Details}\label{app:simulation details}

\subsection{Synthetic Data via Wavelets}\label{app:synthetic data wavelet}

To empirically validate the bounds derived in Proposition \ref{thm: KL ratio} and \ref{thm: Delta KL}, we require data points that satisfy the decomposition:
\begin{align}\label{eq: additive}
    \boldsymbol{x}_h = \boldsymbol{x}_l + \boldsymbol{x}_r.
\end{align}
We construct such data using the Discrete Wavelet Transform (DWT). The DWT provides an exact, non-redundant representation of a signal using a finite set of coefficients and enables lossless reconstruction via its inverse transform \citep{mallat_wavelet_2009}. It also yields a multi-resolution decomposition with basis functions localized in both time and frequency.

Given a signal $\boldsymbol{x} \in \mathbb{R}^s$, the DWT produces a set of coefficients $\{a_J, d_J, d_{J-1}, \dots, d_1\}$, where $a_J$ denotes the approximation coefficients capturing low-frequency content, and $d_j$ denotes the detail coefficients capturing progressively higher frequency components. We construct a low-resolution signal $\boldsymbol{x}_l$ by applying the inverse DWT using only a subset of coefficients (e.g., $a_J$ and selected $d_j$). The residual component $\boldsymbol{x}_r$ is obtained by applying the inverse DWT to the complementary set of detail coefficients. Since the inverse DWT is linear and the coefficient sets form a disjoint partition, their reconstructions add exactly to the original signal, yielding \eqref{eq: additive}. As PCA is also a linear transform, the subsequent projection keeps the additive decomposition. 

\begin{figure}[t]
    \centering
    \includegraphics{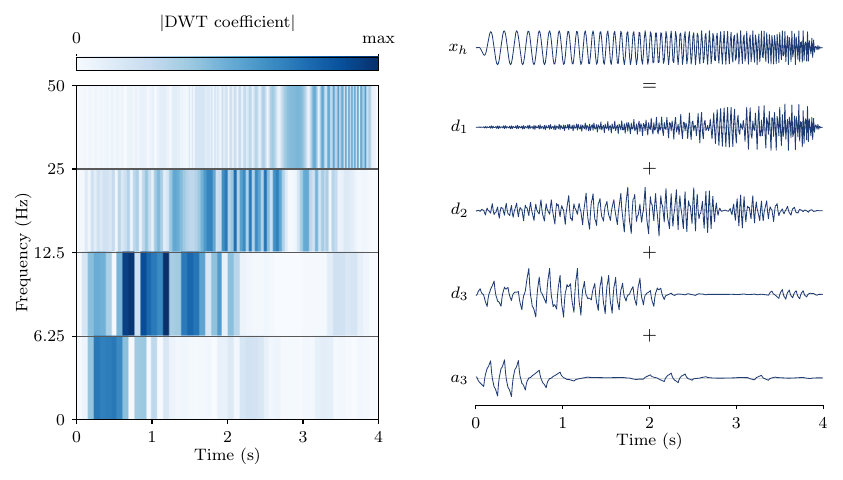}
    \caption{Wavelet construction of the synthetic data. \textbf{Left}: magnitude of the
    db2 DWT coefficients of $\boldsymbol{x}_h$.
    \textbf{Right}: $\boldsymbol{x}_h$ and the reconstruction from each band in isolation. The
    coefficient sets are disjoint and the inverse DWT is linear, so the four
    reconstructions sum exactly to $\boldsymbol{x}_h$, which is the decomposition in
    \eqref{eq: additive}.}
    \label{fig:wavelet-decomposition}
\end{figure}

Figure \ref{fig:wavelet-decomposition} illustrates this construction. On the left we have a scalogram of a synthetic signal, and the right shows exactly what parts of the time series belong to the respective levels in the scalogram. 

\begin{figure}[t]
\centering
\includegraphics{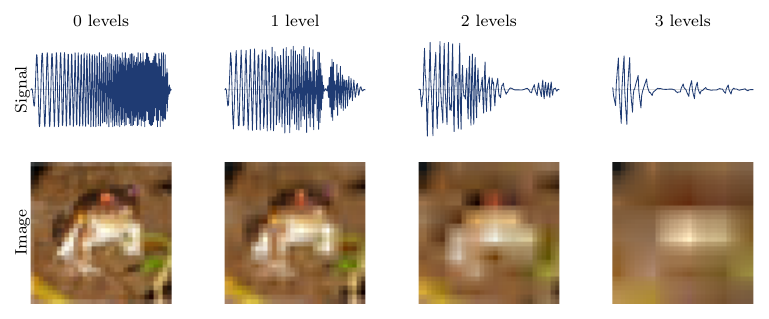}
\caption{Effect of zeroing the $k$ finest detail bands of the db2 DWT, for
$k = 0,\dots,3$, on the synthetic 1D signal (top) and on a CIFAR-10 image
(bottom). This is the operation used to construct the low-resolution inputs
$\boldsymbol{x}_l$.}
\label{fig:wavelet-removal}
\end{figure}

Figure \ref{fig:wavelet-removal} demonstrates what the removal of a different number of levels has of impact on a synthetic time series and a CIFAR10 image. 

\subsection{Exact Values of Bounds}\label{app:simulation bounds}

Table \ref{tab:app simulations} contains the actual KL values, variance approximations and bounds. 
\begin{table}[h]
    \centering
    \caption{Exact values of KL-ratio, KL-diff, and the bounds as seen in Figure \ref{fig:simulation}. Levels refer to the number of frequency bands removed in the scalogram generated from the DWT. LB = lower bound, UB = upper bound, and Cor refers to the tighter bounds in Corollary \ref{cor:tight bounds}.}
    \footnotesize
    \label{tab:app simulations}
    \begin{tabular}{lccccc}
    \toprule 
    \textbf{\# Levels}& \textbf{KL Ratio} & \textbf{Ratio LB} & \textbf{Ratio UB} & \textbf{Ratio LB (Cor)}\\
    \midrule 

Levels 0 & $1.0 \pm 0.00 \!\cdot\! 10^{0}$ & $1.0 \pm 0.00 \!\cdot\! 10^{0}$ & $1.0 \pm 0.00 \!\cdot\! 10^{0}$ & $1.0 \pm 0.00 \!\cdot\! 10^{0}$ \\
Levels 1 & $1.01 \pm 1.10 \!\cdot\! 10^{-4}$ & $0.98 \pm 9.00 \!\cdot\! 10^{-5}$ & $1.02 \pm 9.00 \!\cdot\! 10^{-5}$ & $1.00 \pm 0.00 \!\cdot\! 10^{0}$ \\
Levels 2 & $1.04 \pm 2.00 \!\cdot\! 10^{-4}$ & $0.96 \pm 1.80 \!\cdot\! 10^{-4}$ & $1.05 \pm 1.80 \!\cdot\! 10^{-4}$ & $1.00 \pm 0.00 \!\cdot\! 10^{0}$ \\
Levels 3 & $1.13 \pm 4.30 \!\cdot\! 10^{-4}$ & $0.81 \pm 8.20 \!\cdot\! 10^{-4}$ & $1.21 \pm 1.01 \!\cdot\! 10^{-3}$ & $1.01 \pm 9.00 \!\cdot\! 10^{-5}$ \\
\midrule 
& \textbf{KL Diff} & \textbf{Diff LB} & \textbf{Diff UB} & \textbf{Diff LB (Cor)
}\\
\midrule 
  0 & $0.00 \!\cdot\! 10^{0} \pm 0.00 \!\cdot\! 10^{0}$ & $0.00 \!\cdot\! 10^{0} \pm 0.00 \!\cdot\! 10^{0}$ & $0.00 \!\cdot\! 10^{0} \pm 0.00 \!\cdot\! 10^{0}$ & $0.00 \!\cdot\! 10^{0} \pm 0.00 \!\cdot\! 10^{0}$ \\
 1 & $1.80 \!\cdot\! 10^{-4} \pm 0.00 \!\cdot\! 10^{0}$ & $-2.70 \!\cdot\! 10^{-4} \pm 1.00 \!\cdot\! 10^{-5}$ & $2.70 \!\cdot\! 10^{-4} \pm 1.00 \!\cdot\! 10^{-5}$ & $0.00 \!\cdot\! 10^{0} \pm 0.00 \!\cdot\! 10^{0}$ \\
 2 & $6.00 \!\cdot\! 10^{-4} \pm 1.00 \!\cdot\! 10^{-5}$ & $-7.60 \!\cdot\! 10^{-4} \pm 1.00 \!\cdot\! 10^{-5}$ & $7.80 \!\cdot\! 10^{-4} \pm 1.00 \!\cdot\! 10^{-5}$ & $1.00 \!\cdot\! 10^{-5} \pm 0.00 \!\cdot\! 10^{0}$ \\
 3 & $1.84 \!\cdot\! 10^{-3} \pm 2.00 \!\cdot\! 10^{-5}$ & $-3.09 \!\cdot\! 10^{-3} \pm 6.00 \!\cdot\! 10^{-5}$ & $3.40 \!\cdot\! 10^{-3} \pm 6.00 \!\cdot\! 10^{-5}$ & $1.60 \!\cdot\! 10^{-4} \pm 0.00 \!\cdot\! 10^{0}$ \\

\bottomrule 
    \end{tabular}
\end{table}

\subsection{Approximation Accuracy and Bound Tightness}\label{app:tightness of bounds}
We further investigate the accuracy of the approximation in Proposition \ref{cor:var fraction} and the tightness of propositions \ref{thm: KL ratio}. 

\begin{figure}[h]
    \centering
    \includegraphics[trim=0 0.2cm 0 0.2cm, clip=true]{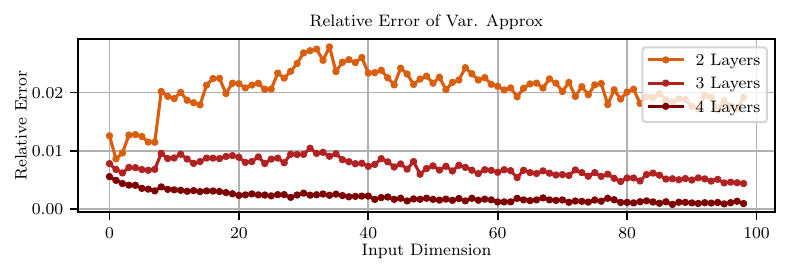}
    \caption{Relative error of the variance approximation in Proposition \ref{cor:var fraction} as a function of input dimension and model size. Three levels of the scalogram are removed from the low-resolution datapoint.}
    \label{fig:var approx rel error}
\end{figure}

\paragraph{Accuracy of Variance Approximation}
We first study the accuracy of the variance approximation in Proposition~\ref{cor:var fraction}. Instead of projecting the synthetic data into a fixed 10-dimensional space, we vary the projected input dimension from 2 to 100. For each input dimension, we compute the relative error
\begin{align}
    e_r = \frac{\mathrm{KL}_h/\mathrm{KL}_l - \mathrm{Var}[\ell(\boldsymbol{\theta}, \boldsymbol{x}_h)] / \mathrm{Var}[\ell(\boldsymbol{\theta}, \boldsymbol{x}_l)]}{\mathrm{KL}_h/\mathrm{KL}_l},
\end{align}
where the low-resolution input has had three wavelet bands removed. We evaluate this quantity for models with 2, 3, and 4 hidden layers. The results can be seen in Figure \ref{fig:var approx rel error}. The relative error decreases as model size increases, and remains below 3\% even for the 2-layer model. This indicates that the variance approximation provides a good description of the KL-ratio. 

\begin{figure}
    \centering
    \begin{subfigure}[b]{1.0\textwidth}
        \includegraphics[trim=0 0.2cm 0 0.2cm, clip=true, width=0.9\textwidth]{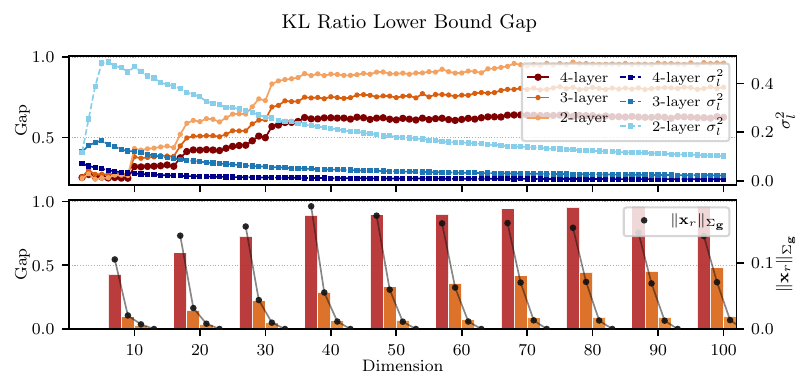}
    \end{subfigure}
    \begin{subfigure}[b]{1.0\textwidth}
        \includegraphics[trim=0 0.2cm 0 0.2cm, clip=true, width=0.9\textwidth]{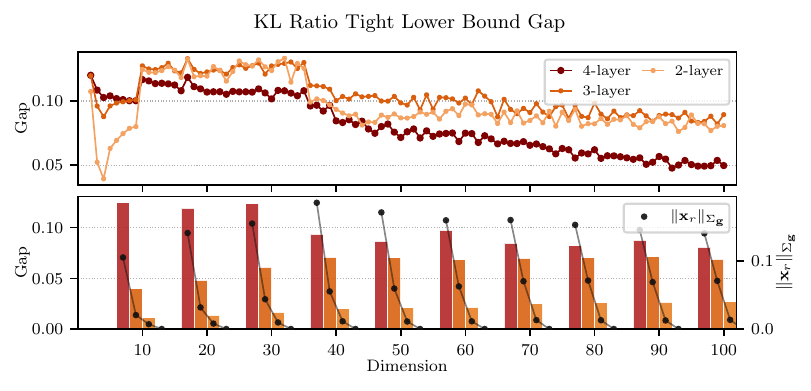}
    \end{subfigure}
        \begin{subfigure}[b]{1.0\textwidth}
        \includegraphics[trim=0 0.2cm 0 0.2cm, clip=true, width=0.9\textwidth]{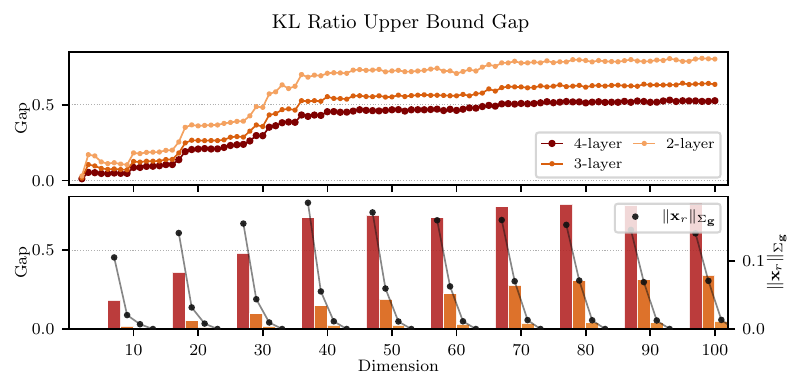}
    \end{subfigure}
    \caption{Tightness of the KL-ratio bounds from Proposition~\ref{thm: KL ratio} and Corollary~\ref{cor:tight bounds} under varying projected input dimension and removed frequency content. Top panels: gap between the true KL-ratio and each bound for 2-, 3-, and 4-layer networks after removing three wavelet bands. Bottom panels: corresponding bound gaps for a 2-layer model under different levels of removed frequency content, with the four bars representing, from left to right, the removal of three, two, one, and zero bands. The directional residual norm $\|x_r\|_{\Sigma_g}$ is overlaid on the secondary axis. The bounds become looser as more residual energy is removed.}
    \label{fig:tightness assessment ratio}
\end{figure}

\paragraph{Tightness of the KL-ratio bounds.}
To study when the bounds in Proposition~\ref{thm: KL ratio} and Corollary~\ref{cor:tight bounds} are tight, we define the gap between the true KL-ratio and a bound as our measure of tightness. For lower bounds, this is the difference between the true value and the bound; for upper bounds, we report the absolute difference. Figure~\ref{fig:tightness assessment ratio} shows this analysis for projected input dimensions from 2 to 100 and for networks with 2, 3, and 4 hidden layers.

In the top panels, three wavelet bands are removed and the resulting gap is plotted as a function of input dimension. The corresponding low-resolution loss variance $\sigma_l^2$ is overlaid, since it appears in all KL-ratio bounds. In the bottom panels, we fix a 2-layer network and vary the number of removed frequency bands. The bars show the corresponding bound gaps, while the directional residual norm $\|\boldsymbol{x}_r\|_{\Sigma_g}$ is shown on the secondary axis. The results show that the bounds become looser as more residual information is removed, and tighter as model size increases.

\subsection{Toy Example}\label{app:toy example}
\begin{figure}[h]
    \centering
    \begin{subfigure}[c]{0.32\textwidth}
        \includegraphics[trim=0 0.2cm 0 0.2cm, clip=true]{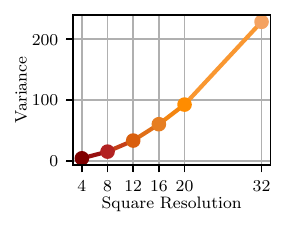}
    \end{subfigure}%
    \hfill 
    \begin{subfigure}[c]{0.67\textwidth}
        \includegraphics[trim=0 0.2cm 0 0.2cm, clip=true]{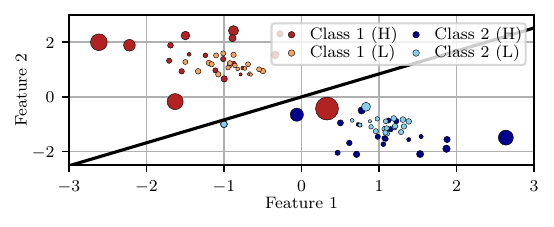}
    \end{subfigure}
    \caption{\textbf{Left}: Variance of low-dimensional representations as a function of image resolution. Computed from PCA projections of progressively downsampled images from CIFAR10. \textbf{Right}: Synthetic two-class data separated using linear discriminant analysis. Marker size given by the magnitude of the change in model parameters when including a datapoint in the training set. The black line is the decision boundary given by training on all points. }
    \label{fig:toy example}
\end{figure}

To illustrate when a high-resolution datapoint can be more informative than its low-resolution counterpart, we consider a simple toy example. The left panel of Figure \ref{fig:toy example} shows that progressively downsampled CIFAR10 representations occupy a more concentrated region after PCA projection. Motivated by this assumption, we construct a two-class synthetic dataset in which low-resolution points have one eighth of the variance of high-resolution points. We fit linear discriminant analysis \citep{fisher_lda_1936} and scale each point by the parameter change induced by its inclusion in the training set. The high-resolution points extending beyond the support of the low-resolution distribution generally produce larger parameter changes, illustrating how information discarded by downsampling can govern datapoint influence. 

\section{Experimental Details}\label{app:training details}
\subsection{Training Procedure}
We process minibatches proportionally to the data mixture: i.e. if the high-resolution ratio is $80\%$, four high-resolution batches are processed for every low-resolution batch. Training uses the AdamW optimiser with constant weight decay of $0.05$, an initial learning rate of $5 \times 10^{-4}$, and a learning-rate schedule consisting of 20 warm-up epochs followed by cosine annealing. Experiments run for $100$ epochs in total. Gradients are clipped to a maximum norm of $1.0$. Experiments on CIFAR are run with a batchsize of 128, 64 for AudioMNIST and 32 for Fed-ISIC-2019. 

Data augmentation for the CIFAR and Fed-ISIC-2019 datasets consists of random crops, horizontal flips, random erasing, and colour jitter. For AudioMNIST, we use random time shifts, random erasing, and magnitude warping. For both CIFAR and AudioMNIST, CutMix and Mixup are applied with probability $0.5$, except for low-resolutions sizes that constitute less than $10\%$ of the high-resolution size. CutMix and Mixup are turned off for the Fed-ISIC-2019 dataset, due to the high class imbalance. 

\subsection{Experimental Setup}
For the transformer-based experiments, we modify the ViT architecture from~\cite{yoshioka-Visiontransformerscifar10Training-2024}. For CIFAR datasets and Fed-ISIC-2019 we use a depth of 8, an attention head number of 6. CIFAR use a patch size of 4 and Fed-ISIC-2019 uses a patch size of 14. For AudioMNIST, we use the same ViT but modify it to handle 1D inputs and choose a depth of 4, an attention head number of 4, and a patch size of 8. The network is configured with a combination of fixed sinusoidal PEs and learnable ones. 

CIFAR and Fed-ISIC-2019 in CNN-based experiments are conducted using the ResNet-18 architecture, where the initial max-pooling layer is removed and first kernel size is changed from 7 to 3 to better accommodate smaller input resolutions. For AudioMNIST we modify the convolutional architecture presented in the original paper to include an adaptive pooling layer such that it can accommodate differently sized inputs.

For the Fed-ISIC-2019 dataset $30\%$ low-resolution means that $30\%$ of the total data pool held by the non-requester data holders is donated. 


\section{Additional Empirical Results}\label{app:res audiomnist}
\begin{figure}[h]
    \centering
    \includegraphics[trim=0 0.2cm 0 0.23cm, clip=true]{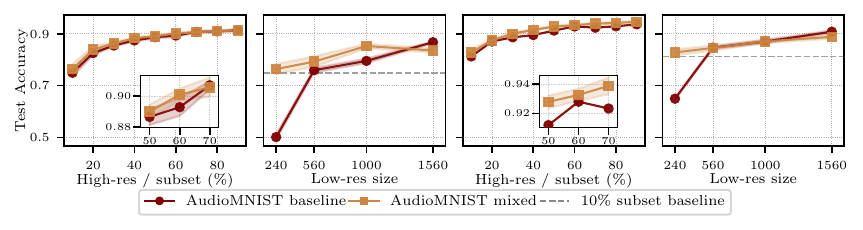}
    \caption{The experiments reported in Figure \ref{fig:cifar results} for AudioMNIST.}
    \label{fig:audiomnist results}
\end{figure}

We perform the same Ratio and Size experiments on AudioMNIST \citep{becker-AudioMNISTExploring-2024} to test whether the findings extend to one-dimensional signals. We use a full sampling rate of $4000$ Hz (downsampled from $8000$ Hz) and low-resolution rates of $\{240, 560, 1000, 1560\}$ Hz, with $240$ Hz used in the Ratio experiments. To avoid overly high accuracy at low subsets, the AudioMNIST training set is reduced from $18,000$ to $5,000$. The results are reported in Figure \ref{fig:audiomnist results}. 

\begin{figure}[h]
    \centering
    \includegraphics{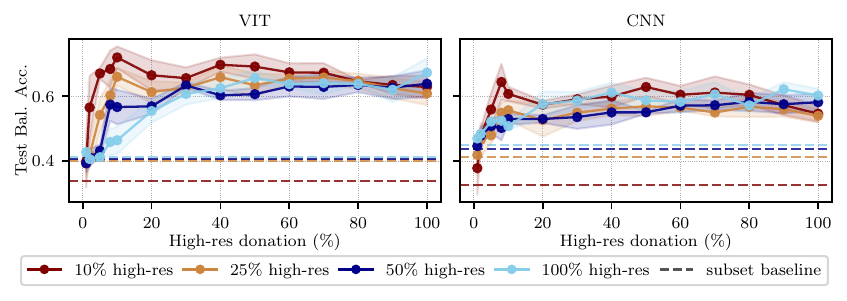}
    \caption{The experiment reported in Figure \ref{fig:fedisic results} but for full-resolution donor data ($224 \times 224$). Everything else is the same including the donation order.}
    \label{fig:fedisic-fullres}
\end{figure}

Figure \ref{fig:fedisic-fullres} shows the Grow experiment of Figure \ref{fig:fedisic results} with donor data added at full resolution, keeping all other settings, including the minority-first donation order, unchanged. Unlike in Figure \ref{fig:fedisic results} the ViT does not decline in balanced accuracy at larger requester subsets; for $100\%$ requester data, the performance increases steadily with the addition of donor data. When training on a $10\%$ requester subset, the performance still peaks at small donations and then decreases somewhat, suggesting that the shifted label distribution of the donors can outweigh the benefit of additional data. Since the donation order is identical in both experiments, class ordering alone cannot explain the decline observed in Figure \ref{fig:fedisic results}. Together with the controlled experiments in Figure \ref{fig:cifar results}, which show that the ViT benefits from low-resolution data without distribution shift, this indicates that the degradation in Figure \ref{fig:fedisic results} arises from the combination of resolution and distribution shift rather than from either alone. 

The CNN shows a similar but weaker early peak in the $10\%$ setting, and is overall more robust to changes in both resolution and label distribution than the ViT.

\section{Ablation Studies}\label{app:ablation studies}

\subsection{Weighting of Low-resolution Loss}\label{sec:training procedure}
Since evaluation is performed exclusively on high-resolution inputs, we anneal the contribution of the low-resolution branch during training while keeping the high-resolution weight fixed at $w_{\mathrm{high}}=1$. We consider two schedules for weighting the low-resolution loss.

\paragraph{1-phase Cosine Schedule}
Let $p=e/E \in [0,1]$ denote the training progress at epoch $e$ out of a total of $E$ epochs. We compute a cosine annealing coefficient $w$, which is used to update the low-resolution loss weight:
\begin{align}\label{eq:schedule 1}
    w = \tfrac{1}{2}\left(1 + \cos(\pi p)\right), \quad
    w_{\mathrm{low}} = \mathrm{ratio}_{\mathrm{low}} \cdot (0.2 + 0.8\,w).
\end{align}
This schedule ensures that low-resolution samples initially contribute in proportion to their dataset share, after which their contribution decays smoothly to one fifth of that share.

\paragraph{2-phase Cosine Schedule}
Let $p = e/E\in [0,1]$  denote the training progress at epoch $e$ out of a total of $E$ epochs. 

The schedule consists of two phases with transition point $p^\star = 0.7$ and floor $a = 0.05$. The low-resolution weight is defined as
\begin{align}\label{eq:schedule 2}
    w_{\mathrm{low}}(p) = \mathrm{ratio}_{\mathrm{low}}\cdot 
\begin{cases}
a + (1-a)\,\tfrac{1}{2}\!\left(1+\cos\!\left(\pi p / p^\star\right)\right),
& 0 \le p \le p^\star, \\[6pt]
a \,\tfrac{1}{2}\!\left(1+\cos\!\left(\pi (p-p^\star)/(1-p^\star)\right)\right),
& p^\star < p \le 1.
\end{cases}
\end{align}

Thus, during the first \(70\%\) of training the low-resolution branch decays smoothly from its data-proportional value $\mathrm{ratio}_{\mathrm{low}}$ to a small floor $a\cdot \mathrm{ratio}_{\mathrm{low}}$, after which it is annealed continuously to zero over the remaining $30\%$. This yields a smooth transition from joint mixed-resolution training to pure high-resolution fine-tuning.

To account for different numbers of high- and low-resolution mini-batches per epoch, we rescale both branch weights as
\begin{align}
     w_{\mathrm{low}} \gets w_{\mathrm{low}} \cdot \frac{0.5}{n_{\mathrm{low}}/(n_{\mathrm{low}}+n_{\mathrm{high}})}, \qquad 
      w_{\mathrm{high}} \gets w_{\mathrm{high}} \cdot \frac{0.5} {n_{\mathrm{high}}/(n_{\mathrm{low}}+n_{\mathrm{high}})}
\end{align}

where \(n_{\mathrm{low}}\) and \(n_{\mathrm{high}}\) denote the numbers of low- and high-resolution mini-batches per epoch. 

All experiments on CIFAR-100 use the 1-phase cosine schedule. On CIFAR-10 and AudioMNIST, CNN-based models performed better under the 2-phase schedule, whereas ViT-based models continued to use the 1-phase schedule. 

Table \ref{tab:low schedule abl} shows an ablation study on the CIFAR10 dataset on selected Ratio experiments. The schedule is compared to weighting the high- and low-resolution loss equally at a value of $1$. The proposed schedules especially make a difference for the lower ratios. For ratios of $70\%$ or $90\%$, high-resolution data already dominates and it thus matters less what the weighting of low-resolution loss is.  
\begin{table}
    \caption{Ablation study on the low-resolution weighting for selected ratio experiments on the CIFAR10 dataset. "Ours" refers to the results seen in \ref{fig:cifar results} with the ViT using the schedule in \eqref{eq:schedule 1} and the CNN the one seen in \eqref{eq:schedule 2}. Equal weighting means both high- and low-resolution losses are weighted with $1$.}
    \label{tab:low schedule abl}
    \centering
\begin{tabular}{ccccc}
    \toprule
    & \multicolumn{2}{c}{\textbf{ViT}} & \multicolumn{2}{c}{\textbf{CNN}} \\
    \cmidrule(r){2-3} \cmidrule(r){4-5}
    Ratio & Ours & Equal Weighting & Ours & Equal Weighting \\
    \midrule
    $10\%$ & $ 73.4 \pm 0.6 $ & $ 69.2 \pm 0.8 $ & $ 77.5 \pm 0.9 $ & $ 75.1 \pm 0.7 $ \\
    $30\%$ & $ 81.5 \pm 0.3 $ & $ 78.9 \pm 0.6 $ & $ 87.9 \pm 0.4 $ & $ 87.1 \pm 0.4 $ \\
    $50\%$ & $ 85.2 \pm 0.4 $ & $ 84.4 \pm 0.3 $ & $ 91.4 \pm 0.3 $ & $ 90.9 \pm 0.4 $ \\
    $70\%$ & $ 87.3 \pm 0.2 $ & $ 87.3 \pm 0.5 $ & $ 93.0 \pm 0.2 $ & $ 92.8 \pm 0.1 $ \\
    $90\%$ & $ 89.2 \pm 0.3 $ & $ 89.1 \pm 0.4 $ & $ 93.8 \pm 0.2 $ & $ 93.8 \pm 0.1 $ \\
    \bottomrule
\end{tabular}

\end{table}

\subsection{Scale Consistency Loss}
\cite{tian-ResFormerScaling-2023} propose a scale-consistency loss to encourage representations learned at different input resolutions to be aligned during training. Concretely, the scale-consistency loss is defined as a smooth-$l1$ loss applied to whitened feature representations. For ViTs, this loss operates on the \texttt{cls}-tokens corresponding to different resolutions:
\begin{align}
    \mathcal{L}_{\text{scale}} =
    \sum_{i=1}^{r-1} \mathcal{L}_{\ell_1}\big(\texttt{cls}_{i+1},\texttt{cls}_i\big),
\end{align}
where $r$ denotes the number of resolutions used during training, and the sum enforces consistency between adjacent resolution levels, with higher resolutions acting as teachers for lower ones.
To assess whether this auxiliary loss is beneficial in our setting, we construct paired high- and low-resolution inputs by first performing a forward pass on the high-resolution samples, then downsampling these samples and passing them through the model again. This yields aligned representations at the two resolutions, to which the scale-consistency loss can be applied. We incorporate this into the loss function for selected Ratio experiments with a weighting coefficient of $0.5$. For CNN architectures, the loss is computed analogously using the post-pooled feature vectors, ensuring comparability across model classes. The results are listed in Table \ref{tab:scale consistency abl}. For both architectures the addition of this auxiliary loss does not yield any significant improvements. 

\begin{table}[h]
    \caption{Ablation study on the scale consistency loss proposed by \cite{tian-ResFormerScaling-2023} on selected Ratio experiments on the CIFAR10 dataset. Loss is added with a coefficient of $0.5$ and applied to the \texttt{cls}-token for the ViT experiments and the post-pooled feature vector for the CNN experiments. "Ours" refer to the results seen in Figure \ref{fig:cifar results}.}
    \label{tab:scale consistency abl}
    \centering
\begin{tabular}{ccccc}
    \toprule
    & \multicolumn{2}{c}{\textbf{ViT}} & \multicolumn{2}{c}{\textbf{CNN}} \\
    \cmidrule(r){2-3} \cmidrule(r){4-5}
    Ratio & Ours & W. Scale Consistency & Ours & W. Scale Consistency \\
    \midrule
    $10\%$ & $ 73.4 \pm 0.6 $ & $ 71.0 \pm 0.5 $ & $ 77.5 \pm 0.9 $ & $ 77.7 \pm 0.7 $ \\
    $30\%$ & $ 81.5 \pm 0.3 $ & $ 79.6 \pm 0.5 $ & $ 87.9 \pm 0.4 $ & $ 87.9 \pm 0.2 $ \\
    $50\%$ & $ 85.2 \pm 0.4 $ & $ 84.1 \pm 0.8 $ & $ 91.4 \pm 0.3 $ & $ 91.3 \pm 0.3 $ \\
    $70\%$ & $ 87.3 \pm 0.2 $ & $ 87.3 \pm 0.4 $ & $ 93.0 \pm 0.2 $ & $ 92.9 \pm 0.1 $ \\
    $90\%$ & $ 89.2 \pm 0.3 $ & $ 89.3 \pm 0.4 $ & $ 93.8 \pm 0.2 $ & $ 93.9 \pm 0.1 $ \\
    \bottomrule
\end{tabular}
\end{table}

\subsection{Regularisation}
\begin{figure}[h]
    \centering
    \includegraphics{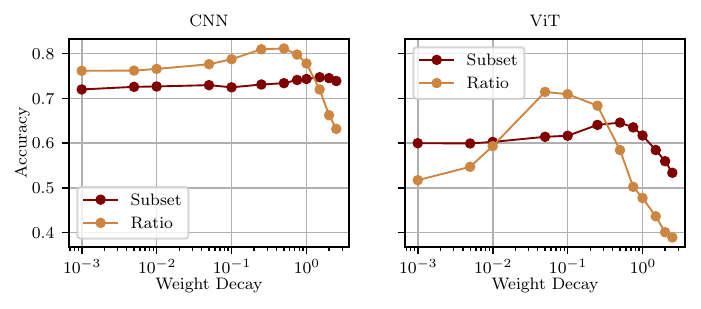}
    \caption{Ablation over weight decay values for the CIFAR10 dataset with $10\%$ high resolution data.  }
    \label{fig:wd ablation}
\end{figure}

To demonstrate that the low-resolution data acts as more than a mere regulariser, we perform some additional experiments on the CIFAR10 dataset with $10\%$ high-resolution data. For both the Ratio and Subset experiments, we sweep the weight decay coefficient over $\lambda \in \{10^{-3}, 5 \cdot 10^{-3}, 10^{-2}, 5 \cdot 10^{-2}, 10^{-1}, 0.25, 0.5, 0.75, 1, 1.5, 2, 2.5\}$. Results for both the ViT and CNN are shown in Figure~\ref{fig:wd ablation}. All results are averaged across five random seeds.

Two observations can be made. First, no weight decay setting in the Subset experiment matches the performance achieved by the best Ratio configuration. This suggests that the gains from incorporating low-resolution data cannot be explained solely by stronger regularisation through weight decay. Second, the ViT is considerably more sensitive to the choice of weight decay than the CNN. Performance improves rapidly as $\lambda$ increases from the smallest values and peaks around $\lambda = 0.05$, after which it gradually declines. This sensitivity indicates that weight decay should ideally be tuned separately for each architecture and experimental setting.

\subsection{Robustness to Corruptions}
\begin{figure}
    \centering
    \includegraphics{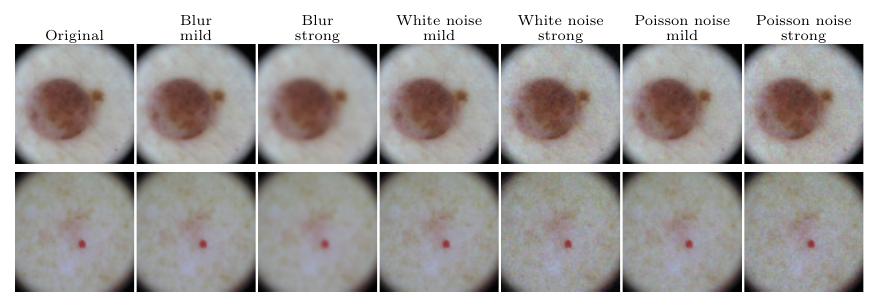}
    \caption{Examples of the acquisition corruptions applied before downsampling. Gaussian blur uses radii of $1.0$ and $2.5$ pixels for the mild and strong settings, respectively. Additive Gaussian noise uses standard deviations corresponding to $3\%$ and $8\%$ of the 8-bit intensity range. Poisson noise uses photon-count scales of $1.0$ and $0.3$, where a smaller scale produces stronger shot noise.}
    \label{fig:corrupt fed}
\end{figure}

The preceding experiments assume that low-resolution observations differ from high-resolution observations only through downsampling. In practice, data collected or transmitted under resource constraints may also be affected by acquisition noise or blur. We therefore evaluate whether the benefit of donated low-resolution data persist under three corruptions: Gaussian blur, additive Gaussian noise, and Poisson noise. Each corruption is applied at mild and strong intensity prior to downsampling. Examples are shown in Figure \ref{fig:corrupt fed}. 

We consider the Fed-ISIC-2019 setting in which the requester trains on $10\%$ of its local data at high resolution and receives all available donor data at low resolution. Table~\ref{tab:fed corrupt} compares the corrupted conditions with both the high-resolution-only Subset baseline and the corresponding clean low-resolution donation setting. Corruption generally reduces the benefit of the donated data relative to clean low-resolution inputs. Nevertheless, every corrupted condition outperforms the high-resolution-only baseline for both architectures. These results indicate that the gains from low-resolution donation are not restricted to ideal downsampling and remain present under moderate acquisition degradation.
\begin{table}
\centering 
    \begin{tabular}{lcc}
\toprule
Condition & ViT & CNN \\
\midrule
Subset baseline (HR only) & 0.339 & 0.326 \\
Clean donation (LR) & 0.566 & 0.484 \\
\midrule
Gaussian noise (mild) & 0.556 & 0.470 \\
Gaussian noise (strong) & 0.550 & 0.470 \\
Gaussian Blur (mild) & 0.581 & 0.475 \\
Gaussian Blur (strong) & 0.561 & 0.480 \\
Poisson noise (mild) & 0.564 & 0.481 \\
Poisson noise (strong) & 0.543 & 0.470 \\
\bottomrule
\end{tabular}
    \caption{Balanced accuracy on Fed-ISIC-2019 when the requester trains on $10\%$ of its local high-resolution data and receives all donor data at low resolution. Corruptions are applied to donor images before downsampling. The high-resolution-only Subset baseline and clean low-resolution donation setting are included for comparison. Despite degradation relative to clean low-resolution data, all corrupted donation settings outperform the Subset baseline.}
    \label{tab:fed corrupt}
\end{table}

\end{document}